\documentclass[letterpaper, 10 pt, conference]{ieeeconf}  

\IEEEoverridecommandlockouts                              

\makeatletter
\let\NAT@parse\undefined
\makeatother

\usepackage{graphics} 
\usepackage{graphicx}
\usepackage{grffile} 
\usepackage{amsmath} 
\usepackage{amssymb}  
\usepackage{color}
\usepackage{bm}
\usepackage{url}
\usepackage{float}
\usepackage{subfig}
\usepackage{wrapfig}

\usepackage[noend]{algpseudocode}
\usepackage{algorithm}
\usepackage{makecell}
\usepackage[square, comma, numbers,sort&compress]{natbib}
\usepackage{multirow}
\let\labelindent\relax
\usepackage{enumitem}
\usepackage[table,dvipsnames]{xcolor}
\usepackage[pagebackref=true,breaklinks=true,colorlinks,bookmarks=false]{hyperref}
\usepackage[noend]{algpseudocode}
\usepackage{algorithm}
\usepackage{times}
\usepackage{textcomp}
\usepackage{booktabs}
\usepackage{threeparttable} 
\usepackage{siunitx}
\usepackage{pifont}
\newcommand{\xmark}{\ding{55}}%
\hypersetup{
linkcolor=BrickRed
,citecolor=Green
,filecolor=Mulberry
,urlcolor=NavyBlue
,menucolor=BrickRed
,runcolor=Mulberry
,linkbordercolor=BrickRed
,citebordercolor=Green
,filebordercolor=Mulberry
,urlbordercolor=NavyBlue
,menubordercolor=BrickRed
,runbordercolor=Mulberry
}

\newcommand\blfootnote[1]{%
  \begingroup
  \renewcommand\thefootnote{}\footnote{#1}%
  \addtocounter{footnote}{-1}%
  \endgroup
}

\title{\LARGE \bf
Simultaneous Navigation and Construction Benchmarking Environments
}

\author{
Wenyu Han, Chen Feng$^{\dagger}$, Haoran Wu, Alexander Gao,\\Armand Jordana, Dong Liu, Lerrel Pinto, Ludovic Righetti\\
\url{https://ai4ce.github.io/SNAC/}
}

\begin{document}

\maketitle
\begin{abstract}
We need intelligent robots for mobile construction, the process of navigating in an environment and modifying its structure according to a geometric design.
In this task, a major robot vision and learning challenge is how to exactly achieve the design without GPS, due to the difficulty caused by the bi-directional coupling of accurate robot localization and navigation together with strategic environment manipulation.
However, many existing robot vision and learning tasks such as visual navigation and robot manipulation address only one of these two coupled aspects.
To stimulate the pursuit of a generic and adaptive solution, we reasonably simplify mobile construction as a partially observable Markov decision process (POMDP) in 1/2/3D grid worlds and benchmark the performance of a handcrafted policy with basic localization and planning, and state-of-the-art deep reinforcement learning (RL) methods.
Our extensive experiments show that the coupling makes this problem very challenging for those methods, and emphasize the need for novel task-specific solutions.
\end{abstract}
\blfootnote{New York University, Brooklyn, NY 11201.
\texttt{\{wenyuhan, cfeng, hw2402, alexandergao, aj2988, ddliu, lerrel, ludovic.righetti\}@nyu.edu}}

\blfootnote{$^{\dagger}$~The corresponding author is Chen Feng. \texttt{cfeng@nyu.edu}}
\section{Introduction}
\label{sec:intro}
\begin{figure}[t!]
\centering
\includegraphics[height=0.17\textwidth]{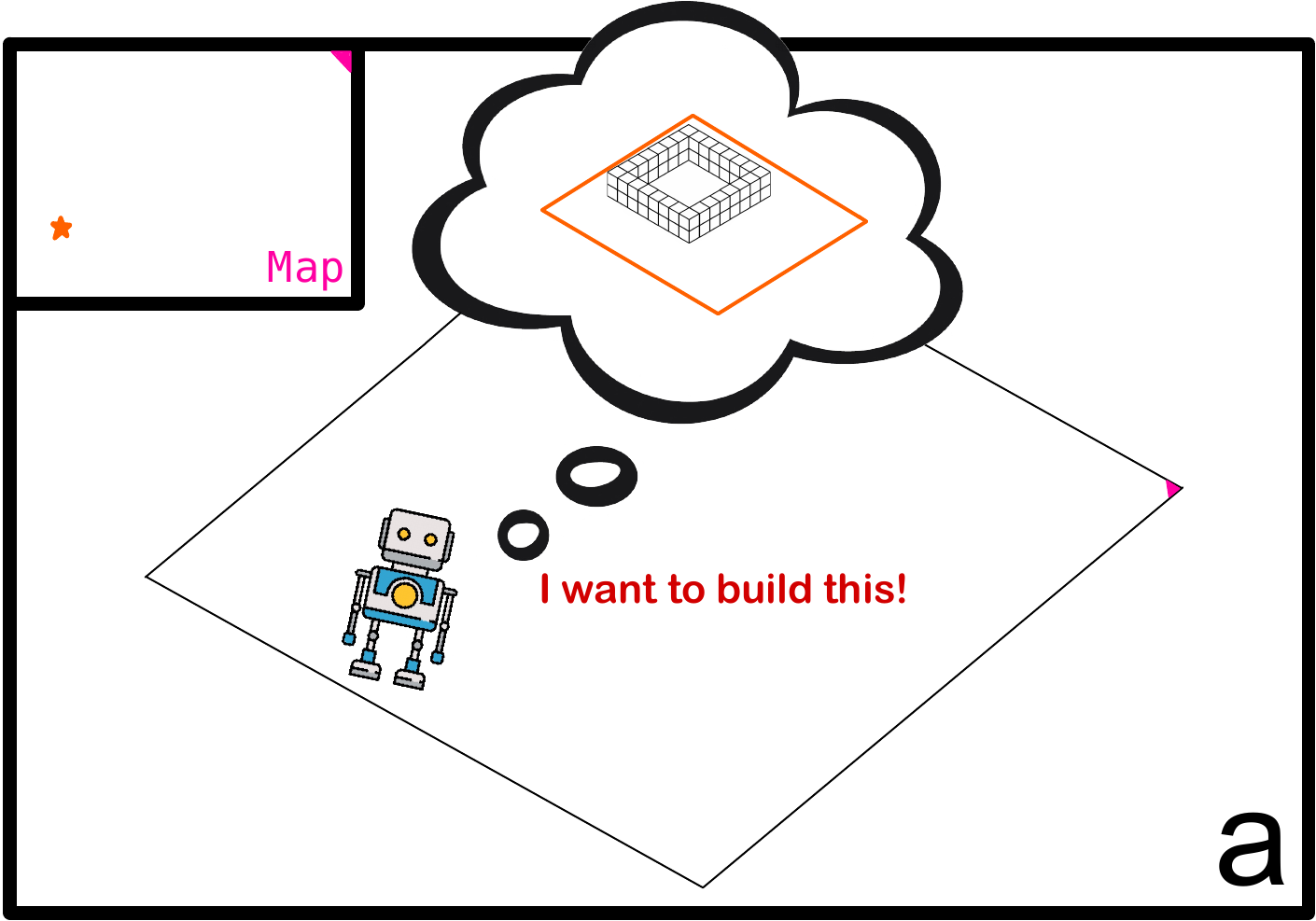}\includegraphics[height=0.17\textwidth]{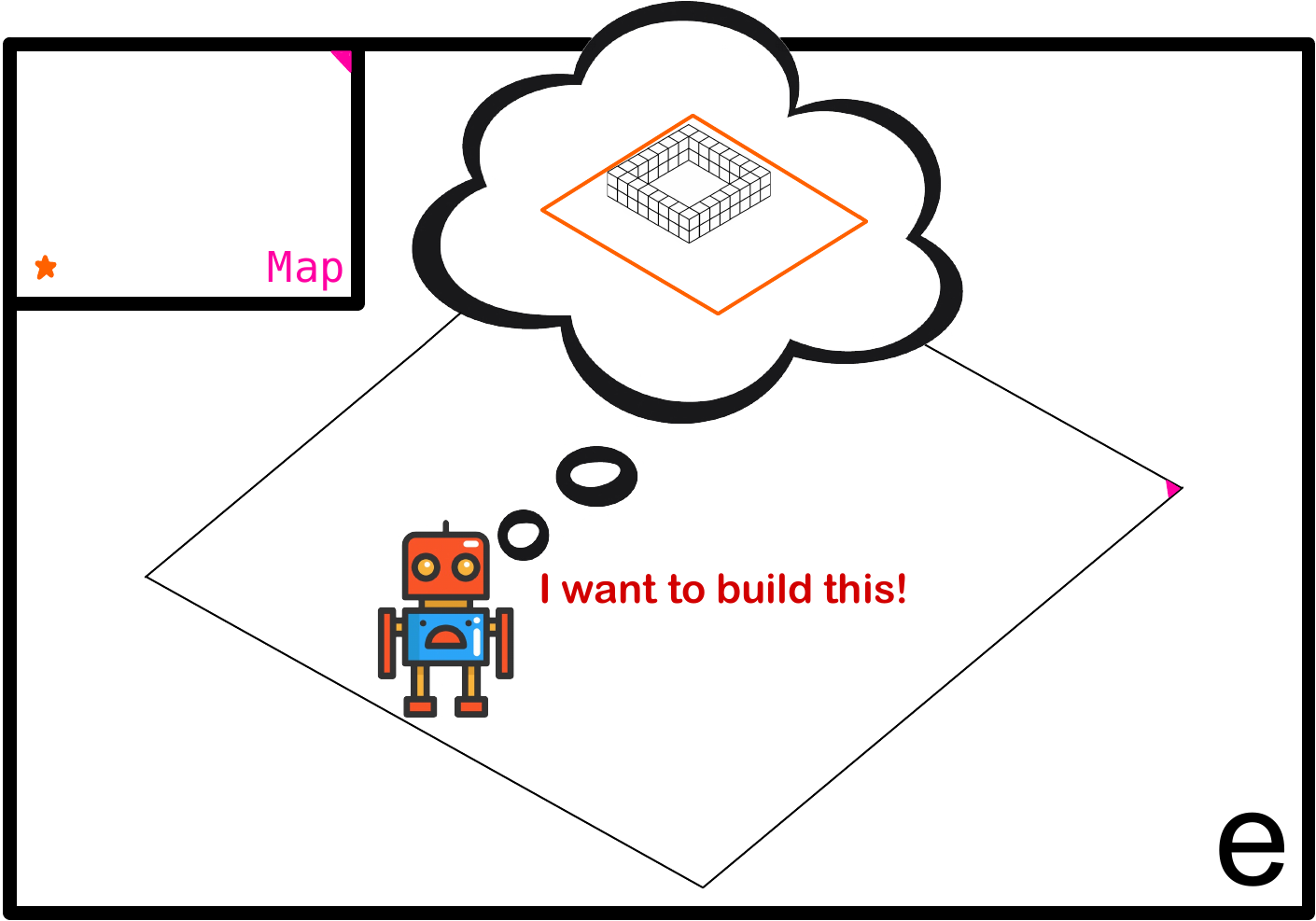}\\
\includegraphics[height=0.164\textwidth]{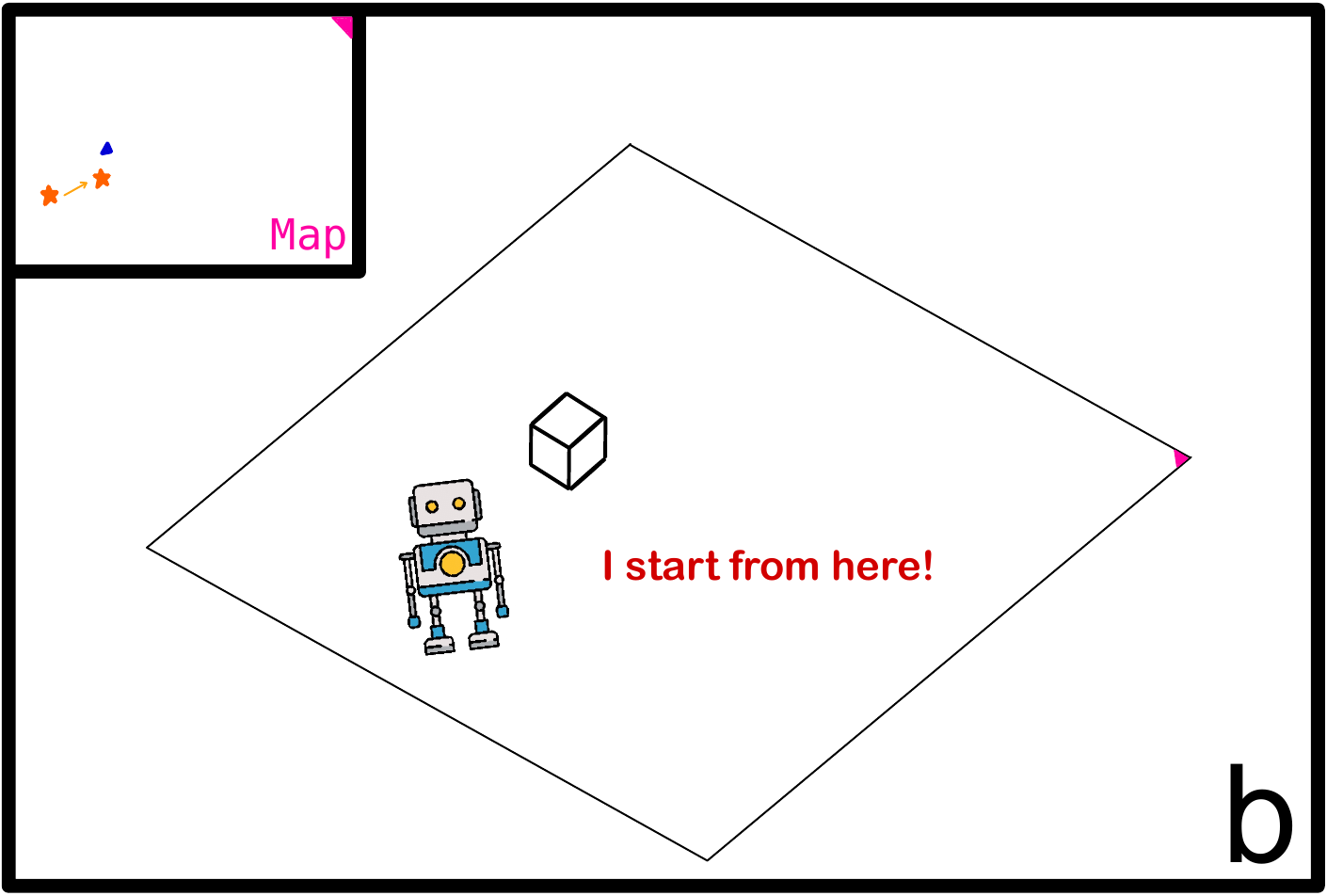}\includegraphics[height=0.164\textwidth]{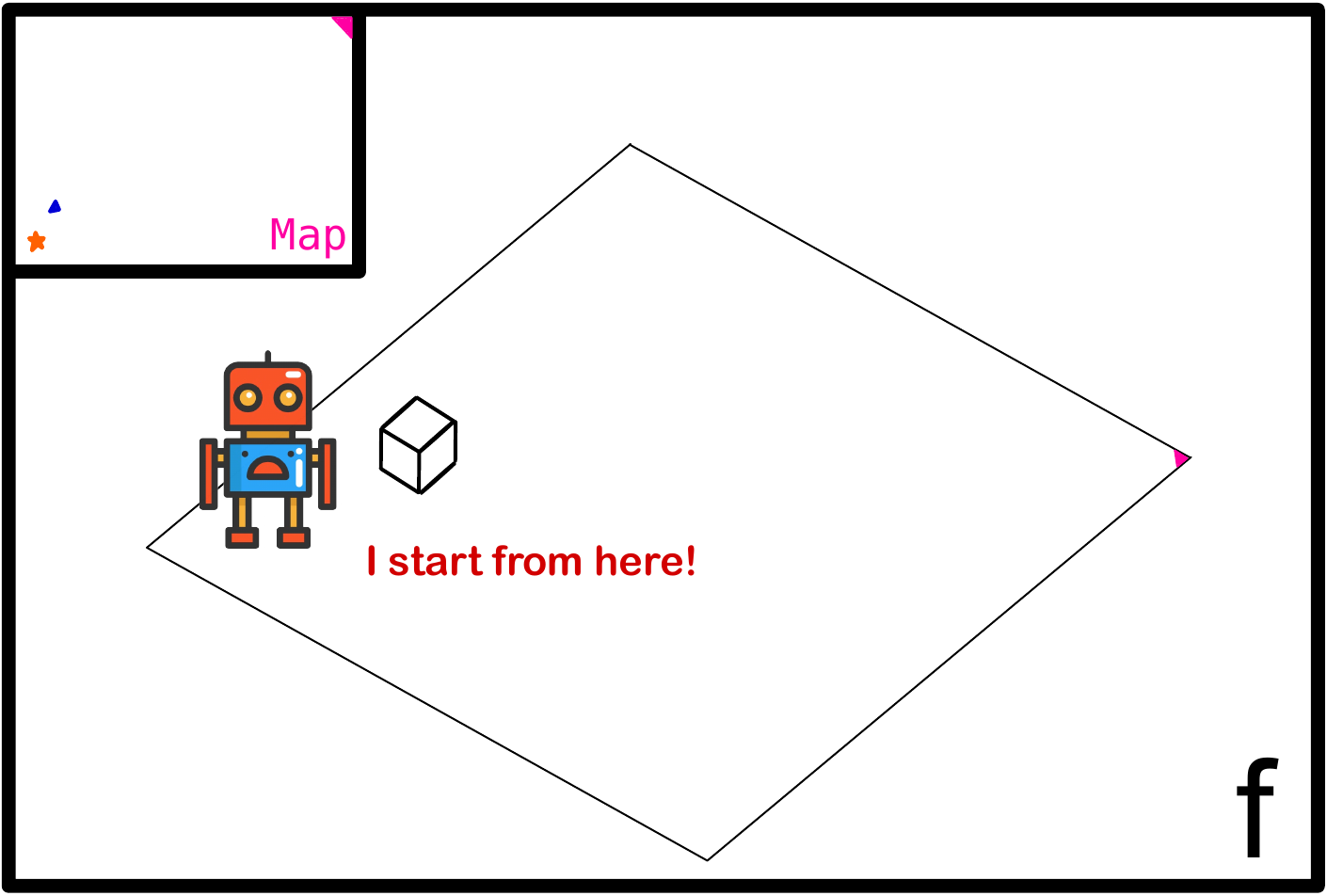}\\
\includegraphics[height=0.164\textwidth]{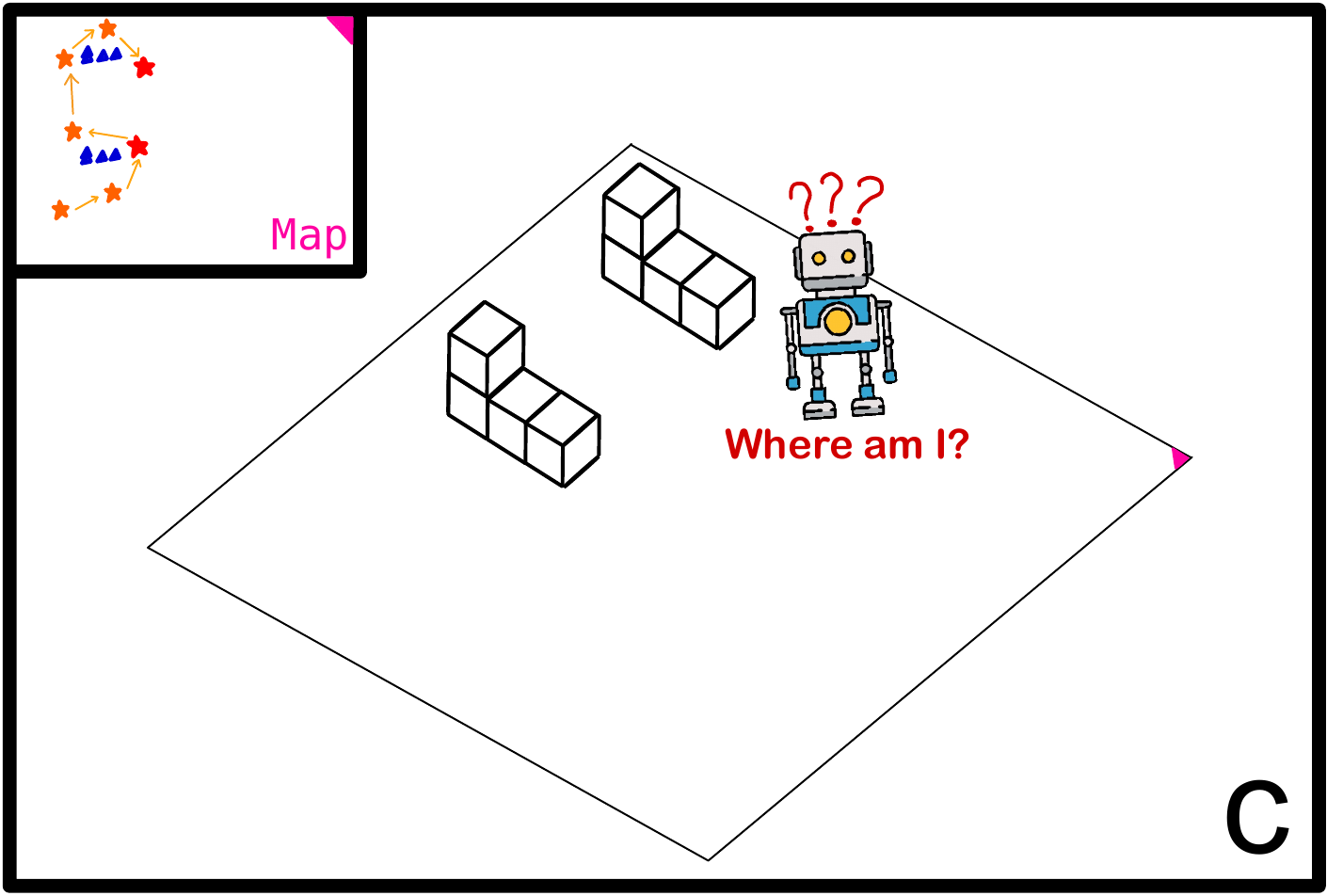}\includegraphics[height=0.164\textwidth]{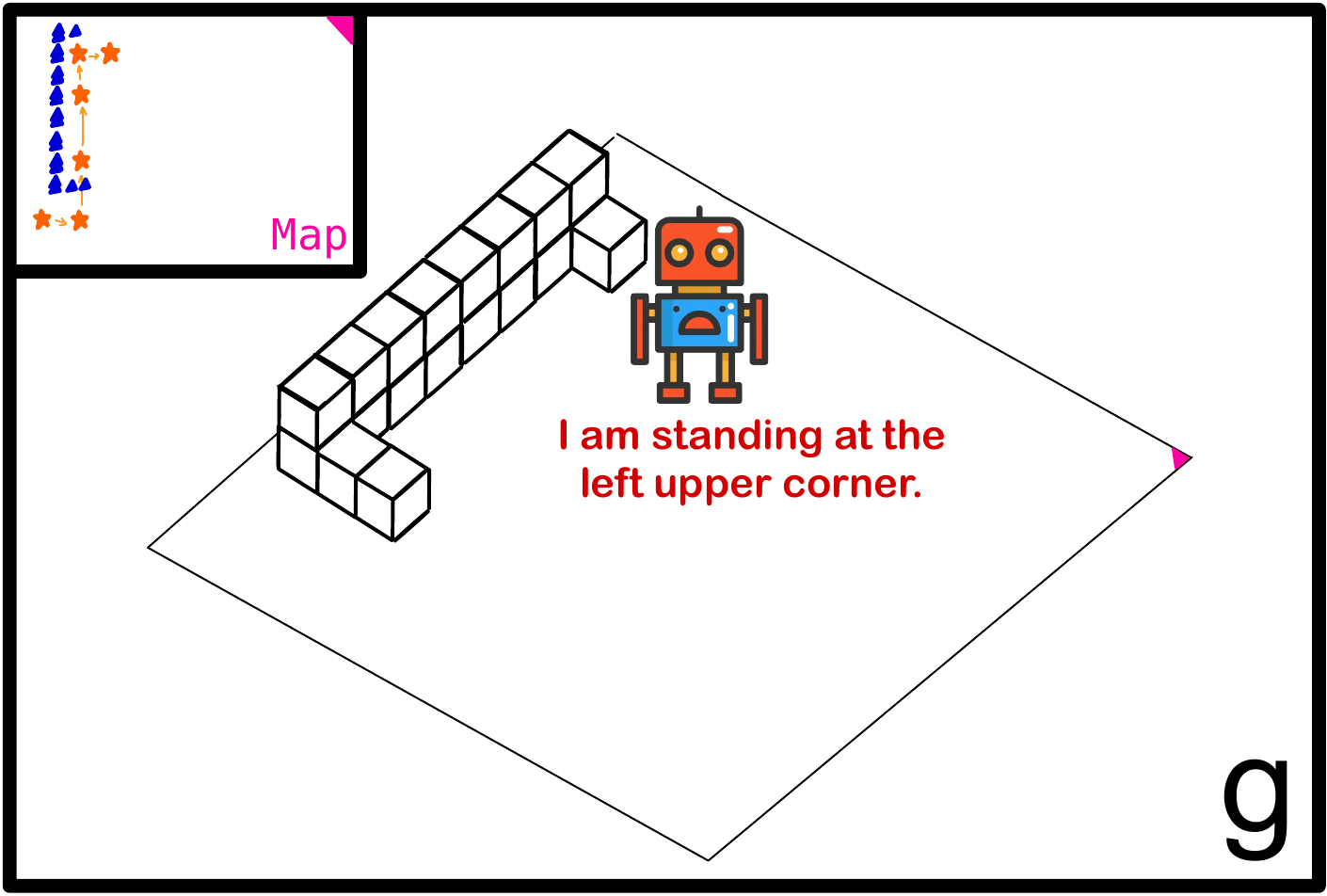}\\
\includegraphics[height=0.164\textwidth]{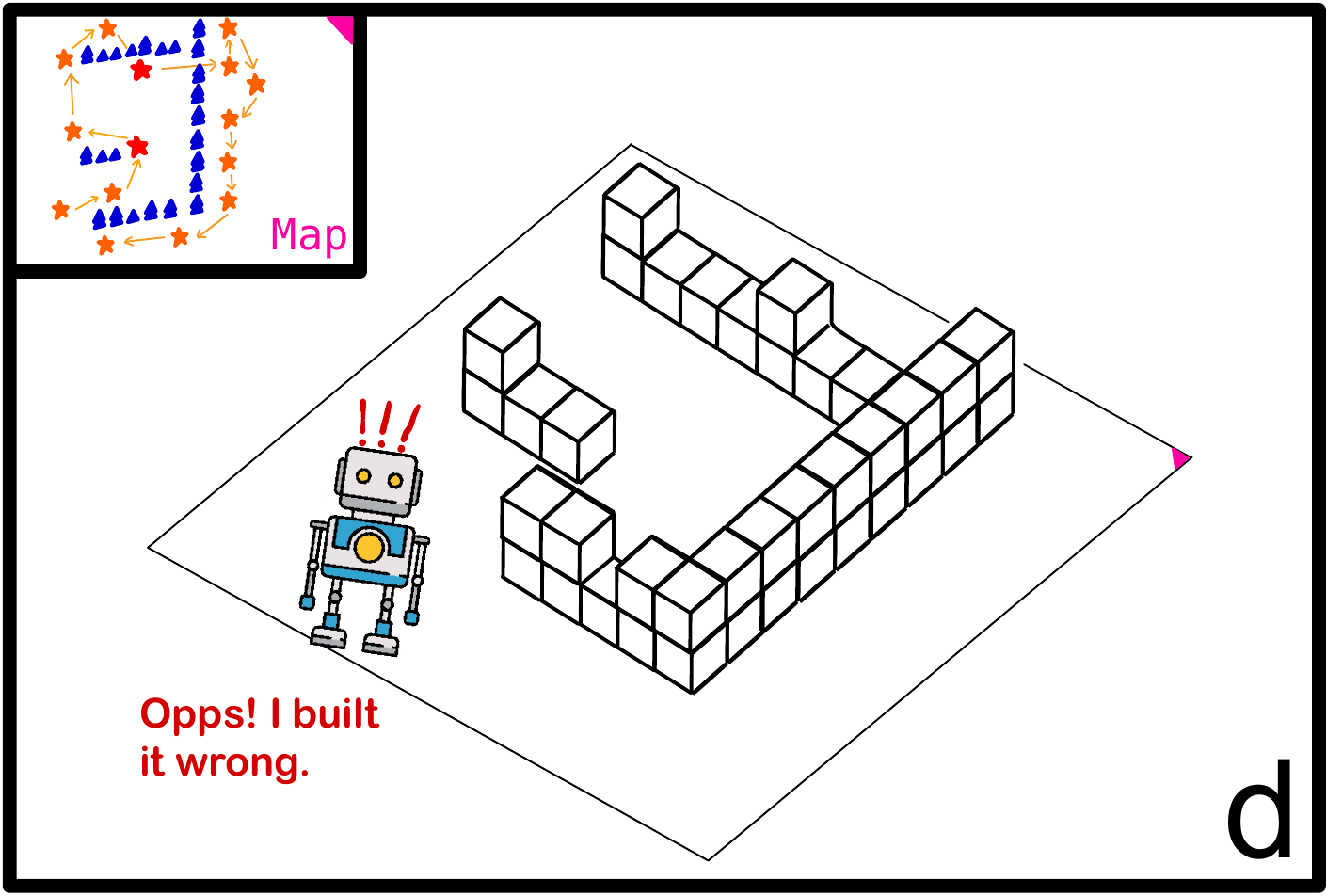}\includegraphics[height=0.164\textwidth]{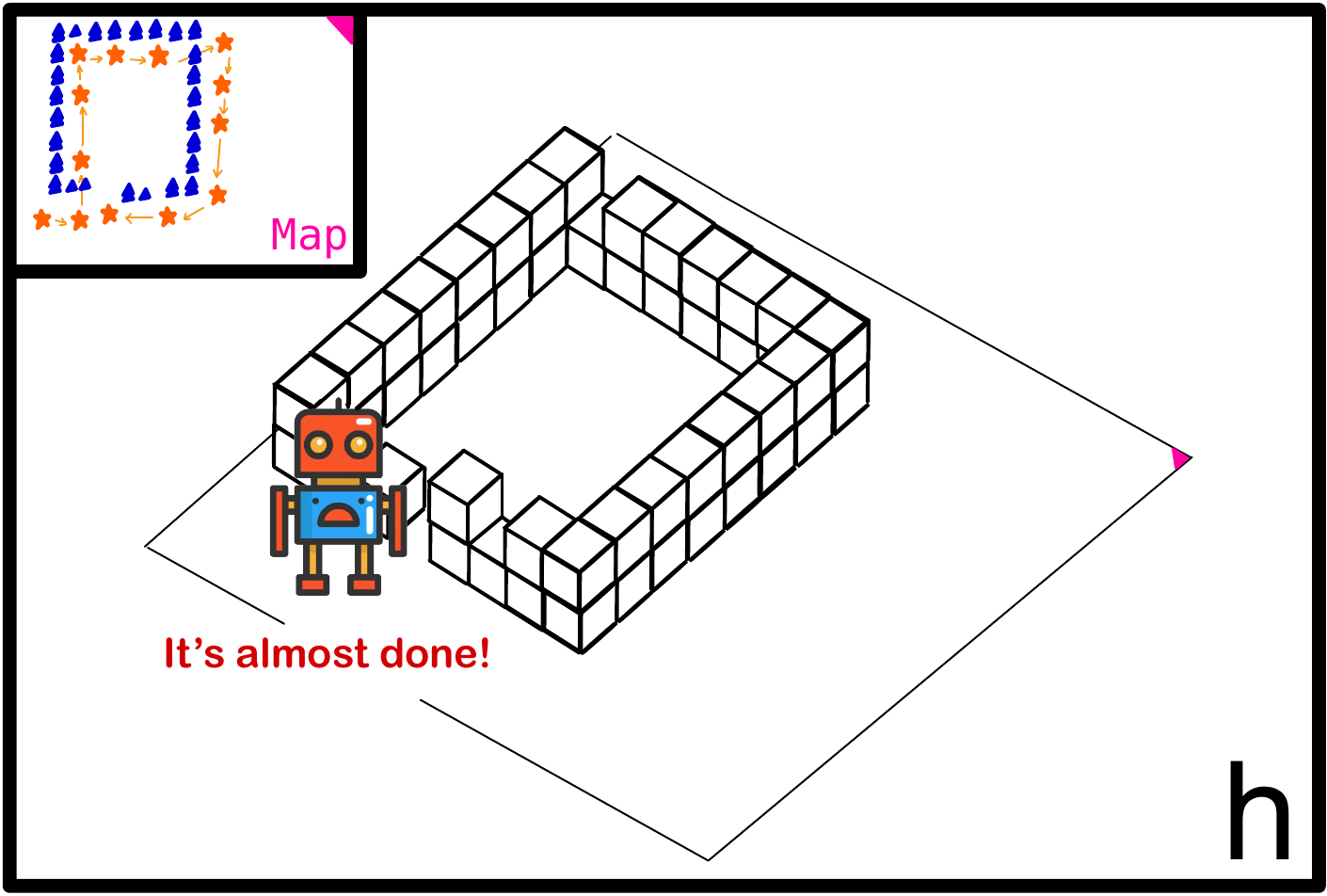}\\
\caption{Illustration of the mobile construction task and its challenges. The robot needs to navigate in an environment (square area) and build a structure according to a geometric design (stacked cubes in a/e). The map shows the ground-truth trajectory (orange) and built structures (blue). The robot in (a-d) has a poor construction planning that does not fully consider the localization, so it loses track of its position because of two similar structures built previously (c). In contrast, a better planning in (e-h) can facilitate localization and lead to better construction.
\label{fig_mobile_construction_example}}
\vspace{-2mm}
\end{figure}

Robotic construction is revived along with AI because of both its growing benefits in time, quality, and labor cost, and the even more exciting role in extraterrestrial exploration. Efficiently achieving this in large scale with flexibility requires robots to have an ability like animal architects (e.g., mound-building termites and burrowing rodents): \textit{mobile construction}, i.e., the process of an agent simultaneously navigating around a site and manipulating its surroundings according to a design.

Engineering mobile construction using existing technologies is difficult. Needless to mention the materials and mechanics problems, a fundamental challenge is \textit{the tight bi-directional coupling of robot localization and long-term planning for environment modification}, which involves core robot vision topics such as simultaneous localization and mapping (SLAM) and structure from motion (SfM). If GPS and techniques alike are not available (often due to occlusions), robots have to rely on SLAM for pose estimation. But mobile construction violates the basic static environment assumption in matured visual SLAM methods, and even challenges SfM methods designed for dynamic scenes~\citep{saputra2018visual}. This means any method addressing this challenge has to take advantage of that bi-directional coupling to strategically modify the environment while efficiently updating a memory of the evolving structure in order to perform accurate localization and construction, as shown in Figure~\ref{fig_mobile_construction_example}.



Deep reinforcement learning (RL) offers another possibility to tackle this challenge, especially given the recent success of deep model-free RL in game playing and robot control. Can we train deep networks to learn a generic and adaptive policy that controls the robot to build calculated localization landmarks as temporary structures which eventually evolve to the designed one?
This calls for a simulation environment for mobile construction. However real-world construction contains more complexities such as the physical dynamics between robots and environments, which are important but less relevant in our context.
Thus, to stay focused, we design a series of mobile construction tasks in 1/2/3D grid worlds with simplified environment dynamics and sensing models, solving which would bring us closer to intelligent mobile construction robots.

To show the reasonably simplified tasks are still non-trivial and challenging, we benchmark the performance of several baselines, including human, handcrafted policy with basic SLAM and planning, and some deep RL algorithms which have achieved state-of-the-art performance in other robotics and RL tasks (see Table~\ref{existed_RL} for comparisons).
Surprisingly, although our tasks may seem similar to other grid/pixelized tasks such as Atari games~\citep{mnih2013playing}, the results reveal many difficulties for those algorithms.


In summary, our contributions include: 
\begin{itemize}[nosep,nolistsep]
\item a suite of novel and easily extensible tasks directly connected to important real-world robot localization and planning problems, which will be released as open-source fast Gym environments (in the supplementary); 
\item a comprehensive benchmark of baseline methods, which demonstrates the unexpected difficulties in these tasks for previously successful algorithms;
\item a detailed ablation study revealing insights about the causes of those difficulties as the coupling explained above, which may call for novel algorithms from this community to solve the problem more effectively.
\end{itemize}

\section{Related Works}
\label{sec:related_works}
\textbf{Mobile construction robots}. Recently we have seen a rising trend of 3D printing using mobile robots all around the globe for construction and manufacturing~\citep{werfel2014designing,jokic2014robotic,ardiny2015construction,nan2015new,marquesmobile,buchli2018digital,zhang2018large,melenbrink2020autonomous}. All of those are carefully engineered systems that either assume some global localization ability or only work for specific scenarios, which restricts their feasibility in large scale. Moreover, none of them address the aforementioned challenges from a theoretical perspective.

\textbf{POMDP solvers}. We model mobile construction tasks by POMDP. An intuitive thought is to assume a perfect knowledge of the transition and observation models (although we do not), and then try them on existing offline or online solvers such as SARSOP~\citep{kurniawati2008sarsop} or POMCP~\citep{silver2010monte}. But due to environment modification, both our state and observation spaces are huge, compared with the existing large POMDP tasks such as in~\cite{wandzel2019multi} with $10^{27}$ states. Our simplest 1D task can easily have a much larger state space ($100^{100}$ for 100 grids with max height 100, detailed in Section~\ref{sec:tasks}) than the Go game ($3^{361}$ states). 
Since it is non-trivial to design a model-based baseline for our tasks, we will mainly focus on benchmarking model-free RL algorithms and leave the investigation of model-based methods for our future work.



\begin{table}[t]
\vspace{3mm}
\centering

\scalebox{0.9}{\begin{tabular}{c|c|c|c|c}
\hline
&  \textbf{Loc} & \textbf{Plan} & \textbf{Env-Mod} &\textbf{Env-Eval} \\ \hline

Robot Manipulation~\citep{fan2018surreal,yang2019replab,labbe2020monte,li2020hrl4in} & \xmark & \checkmark & \checkmark & \xmark \\ \hline 
Robot Locomotion~\citep{duan2016benchmarking} & \xmark & \xmark & \xmark& \xmark  \\ \hline
Visual Navigation~\citep{zhu2017target,gupta2017cognitive,mo2018adobeindoornav,zeng2020survey} & \checkmark& \checkmark & \xmark& \xmark \\ \hline 

Atari~\citep{mnih2013playing} & \xmark & \checkmark/\xmark & \checkmark& \xmark \\\hline 
Minecraft~\citep{oh2016control,guss2019minerl,platanios2020jelly} & \xmark & \checkmark/\xmark & \checkmark &\xmark\\\hline 
First-Person-Shooting~\citep{lample2017playing}  & \checkmark & \xmark & \xmark & \xmark \\\hline 
Real-Time Strategy Games~\citep{synnaeve2016torchcraft,jaderberg2019human} & \checkmark & \checkmark & \checkmark & \xmark \\ \hline 
Physical Reasoning~\cite{bapst2019structured,bakhtin2019phyre} & \xmark & \checkmark & \checkmark & \xmark \\ \hline
\textbf{Mobile Construction (Ours)} & \checkmark& \checkmark & \checkmark& \checkmark 

\\\hline 
\end{tabular}}
\caption{Mobile construction vs. existing robotics and RL tasks. \textbf{Loc}: tasks require robot localization. \textbf{Plan}: tasks require long-term planning. \textbf{Env-Mod}: tasks require environment structure modification. \textbf{Env-Eval}: evaluations are based on the accuracy of environment modifications. The table clearly shows mobile construction is a novel task that exhibit fundamentally different features than typically benchmarked RL tasks, requiring joint efforts of robot localization, planning, and learning.
\label{existed_RL}}
\vspace{-5mm}
\end{table}


\textbf{RL baseline selection}. 
We create baseline methods from state-of-the-art model-free deep RL methods. Since our tasks are in grid worlds similar to many Atari games, e.g., the Pac-Man, our first choice is DQN~\citep{mnih2013playing} which has achieved great success on many Atari tasks with high-dimensional states, benefiting from better Q-learning on representations extracted via deep networks.
Another baseline is Rainbow~\citep{hessel2017rainbow}, combining six extensions on top of the base DQN, achieving superior performance to any of the individual extensions alone.
These include double Q-learning~\citep{NIPS2010_3964}, prioritized experience replay~\citep{schaul2016prioritized}, dueling network architectures~\citep{wang2016dueling}, multi-step TD learning~\citep{sutton1988}, noisy networks~\citep{DBLP:journals/corr/FortunatoAPMOGM17}, and distributional reinforcement learning~\citep{DBLP:journals/corr/BellemareDM17}.

Of course, DQN is sub-optimal for POMDP due to its limited ability to represent latent states from long-term history, which could be critical for mobile construction. To address this issue, we add a baseline using DRQN~\citep{hausknecht2015deep} with a recurrent Q-network. Besides, a sparse reward function may bring additional challenges to our tasks. Hindsight Experience Replay~\citep{andrychowicz2017hindsight} helps off-policy RL learn efficiently from sparse rewards without complex reward engineering, which is combined with DRQN as another baseline.

In addition, we add two actor-critic based baselines. One is Soft Actor-Critic (SAC) for discrete action settings~\citep{christodoulou2019soft}. Compared with the original SAC~\citep{haarnoja2018soft}, SAC-Discrete inherits the sample efficiency and tailors the effectiveness for discrete action space which suits our tasks. The other is Proximal Policy Optimization (PPO)~\citep{schulman2017proximal}. Whereas the standard policy gradient method performs one gradient update per data sample, PPO enables multiple epochs of minibatch updates by a novel objective with clipped probability ratios.

\textbf{Non-learning basline}. Besides the above methods, one may wish to see the performance of more classical approaches such as (either dynamic~\citep{saputra2018visual} or active~\citep{mu2016information}) visual SLAM, even though they are not designed to work in our dynamic environments, nor do they address the planning for better construction accuracy. Therefore, we implement a naive handcrafted policy (pseudo code in the supplementary) with basic localization and planing modules as a non-learning baseline. The localization module borrows the idea from visual SLAM~\citep{saputra2018visual} which relies on finding the common features through successive images to estimate the robot pose. The planing module simply controls the robot to always build at the nearest possible location. 

\section{Mobile Construction in Grid World}\label{sec:tasks}
We formulate a mobile construction task as a 6-tuple POMDP $\langle\mathcal{S},\mathcal{A},\mathcal{T},\mathcal{O},\mathcal{R},D\rangle$, in which a robot is required to accurately create geometric shapes according to a design $D$ in a grid world. The state space $\mathcal{S}$ is represented as $\mathcal{S}=\mathcal{G}\times \mathcal{L}$, where $\mathcal{G}$ is a space of all possible grid state $G$ storing the number of bricks at each grid, and $\mathcal{L}$ is a space of all possible robot locations $l$ in the grid world.
At each time step, the robot takes an action $a\in\mathcal{A}$, either moving around or dropping a brick at or near its location. Moving a robot will change its location according to the \textit{unknown} probabilistic transition model $\mathcal{T}(l'|l,G,a)$. Meanwhile for simplicity, dropping a brick will change the grid state $G$ at or near $l$ \textit{without any uncertainty}. This is reasonable because in mobile construction settings, motion uncertainty is often the key factor to ensure accuracy~\cite{sandy2016}.
The robot can make a local observation $o\in\mathcal{O}$ of $G$ centering around its current location, with a sensing region defined by a half window size $W_s$. We pad constant values outside the grid world boundary to ensure valid observations in $\mathcal{L}$. Specifically, we use the value -1 to distinguish it from empty (=0) or filled ($>0$) grid states. This could help the robot localize itself near the boundary. Finally, the design $D\in\mathcal{G}$ is simply a goal state of the grid world the robot needs to achieve, and $\mathcal{R}(s,a;D)$ is the reward function depending on this design.

\begin{table*}[htp]
\centering

\begin{tabular}{c|c|c|c}
\hline
\textbf{Dimension}            & \textbf{1D} & \textbf{2D} & \textbf{3D} \\ \hline
\textbf{Action} $a$           &  \thead{Move-left/-right,\\ drop-brick at current location}  & \thead{Move-left/-right/-up/-down,\\drop-brick at current location}& \thead{Move-left/-right/-forward/-backward, \\drop-brick on left/right/front/rear side}  \\ \hline
\textbf{Grid state} $G$ &$\mathbb{R}^W$ & ${[0,1]}^{W\times H}$   & $\mathbb{R}^{W\times H}$   \\ \hline
\textbf{Observation} $o$ &  $\mathbb{R}^{2W_s+1}$  &  ${[0,1]}^{(2W_s+1)\times (2W_s+1)}$  &  $\mathbb{R}^{(2W_s+1)\times (2W_s+1)}$  \\ \hline
\textbf{Dynamic} \& \textbf{Static}                &  \checkmark  &  \checkmark  &  \checkmark  \\ \hline
\textbf{Dense} \& \textbf{Sparse}                & \xmark & \checkmark  & \checkmark   \\ \hline
\textbf{Obstacle}             & \xmark  & \xmark   & \checkmark   \\ \hline

\end{tabular}
\caption{Detail setups of each character for 1/2/3D environment. Each character is described in Section~\ref{sec:tasks} and illustrated in Figure~\ref{fig_example}. In 1D environments, the grid state $G\in\mathbb{R}^W$ is a vector, where $W$ is the width of the environment and $o\in\mathbb{R}^{2W_s+1}$ is a vector with size $2W_s+1$. The grid state $G$ for 2D and 3D environment are 2D binary matrix and 2D matrice with width $W$ and height $H$ respectively. We only add obstacle mechanism in 3D environment.}
\label{tab_env_detail}
\vspace{-3mm}
\end{table*}

The aforementioned coupling challenge in localization and planning are reflected via two factors in this setting.
First, the partial observability makes robot localization necessary.
Second, the environment uncertainty in $\mathcal{T}$ simulates real world scenarios where motion control of the mobile robot is imperfect and the odometry is error-prone. We implement this by sampling the robot's moving distance $d$ in each simulation time step from a uniform distribution.

\begin{figure}[htb]
\centering
\vspace{-3mm}
\subfloat[][1D environment] {\includegraphics[width=0.18\textwidth]{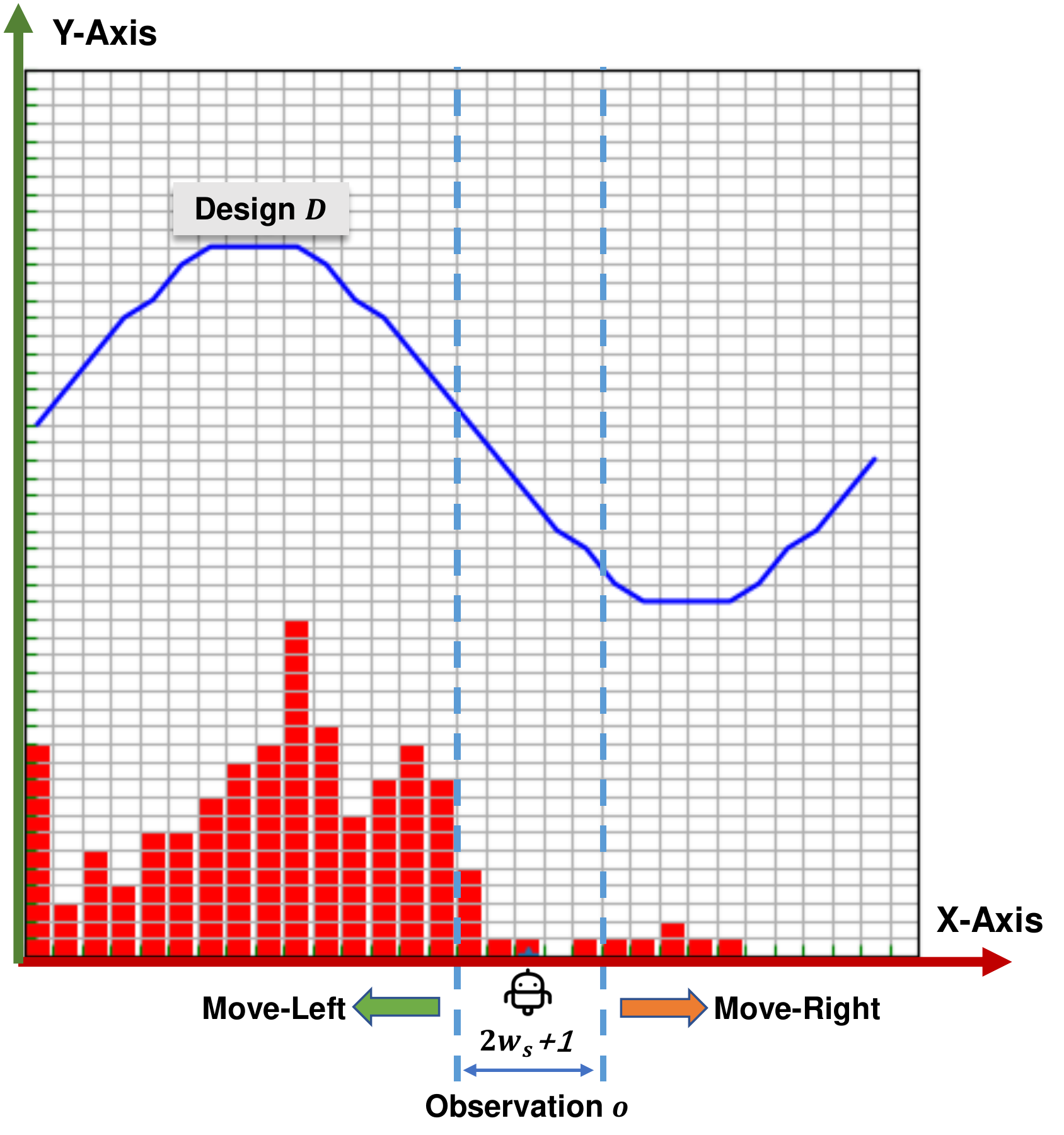}\label{Fig_1D_example}} \hspace{0.2mm}
\subfloat[][2D environment] {\includegraphics[width=0.19\textwidth]{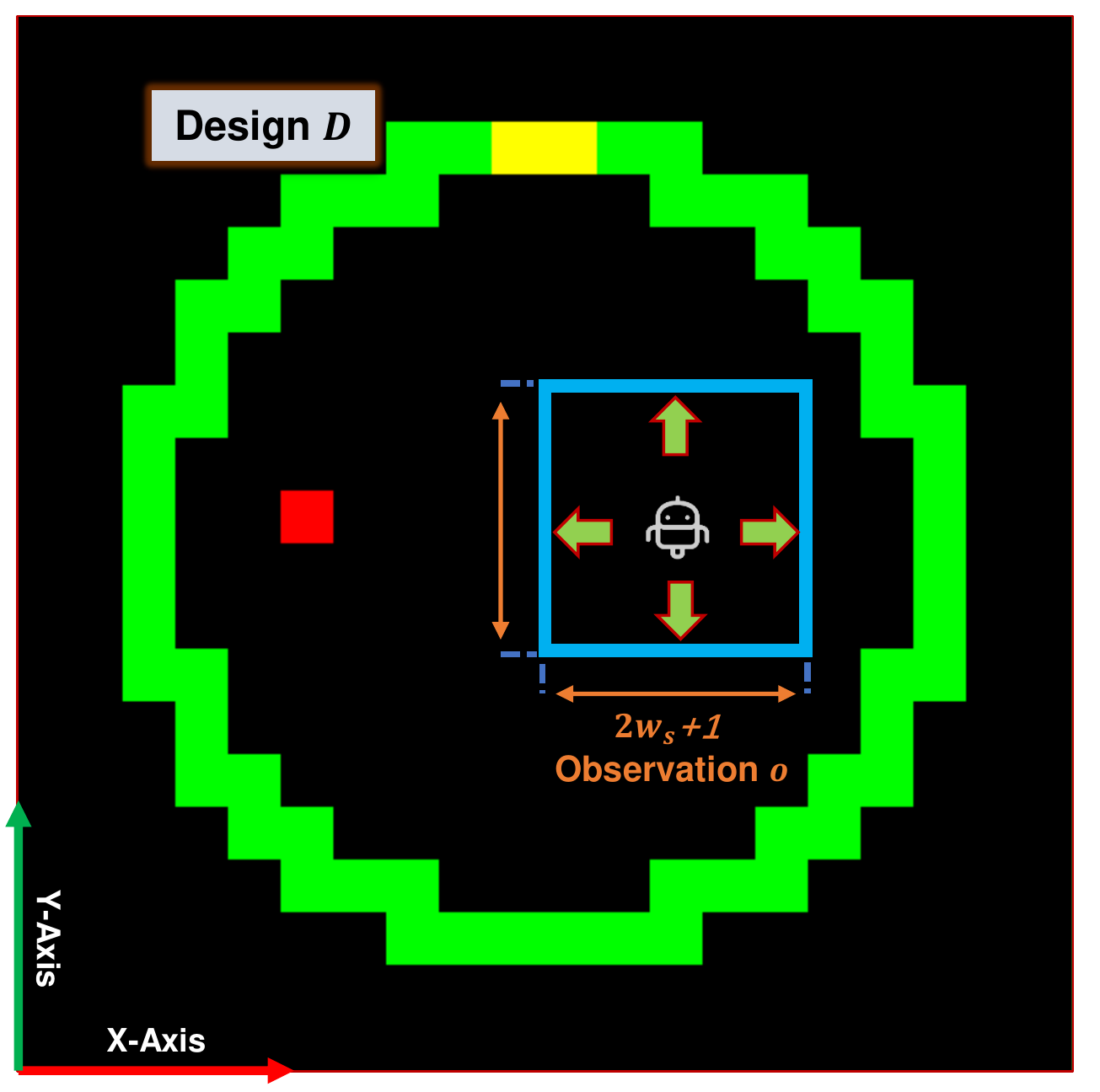}\label{Fig_2D_example}}
\vspace{-2mm}

\subfloat[][3D environment] {\includegraphics[width=0.37\textwidth]{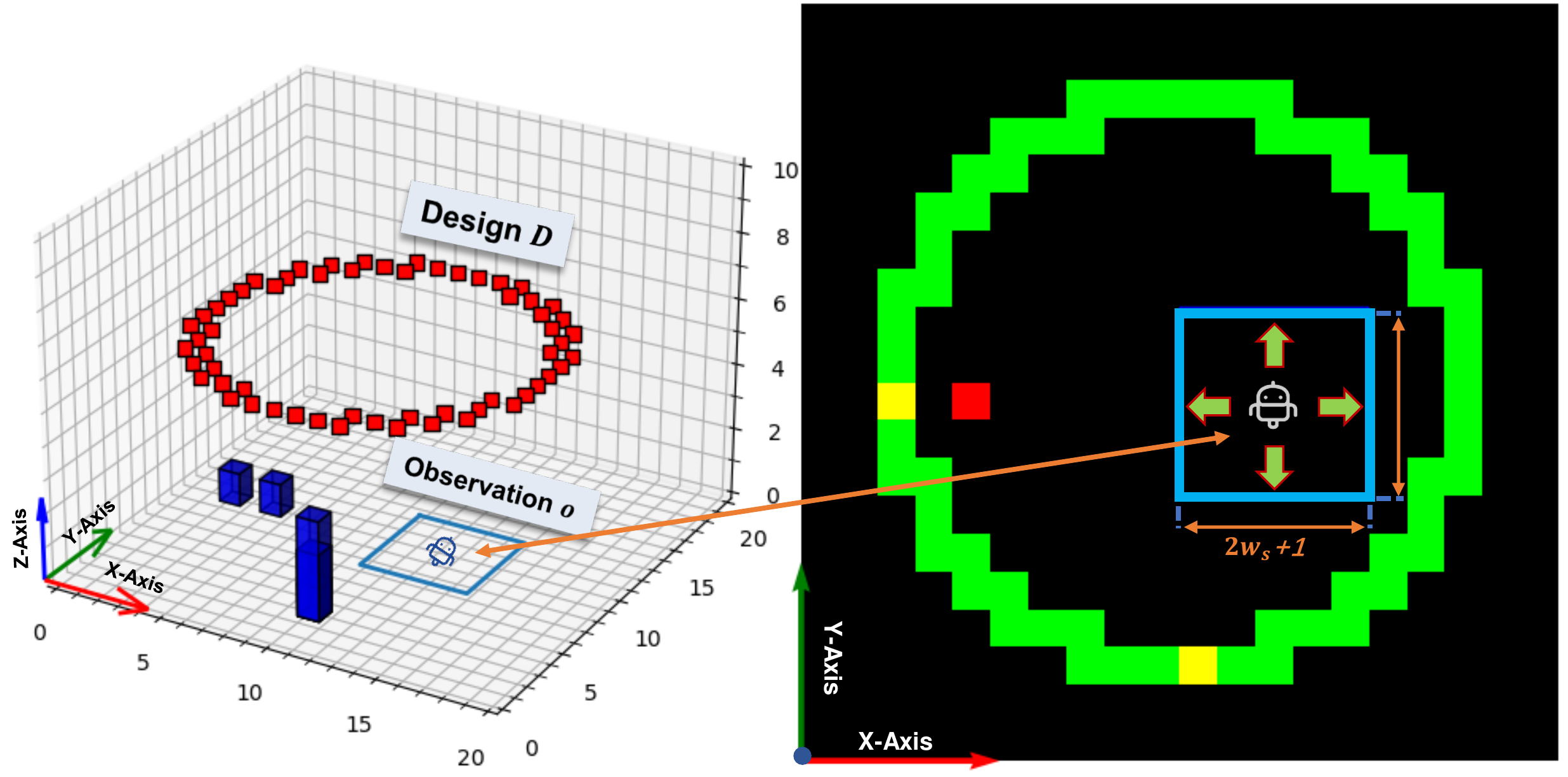}\label{Fig_3D_example}}\\
\vspace{1mm}
\caption{(a) 1D grid world environment: a robot moves along the x-axis and drops bricks (red) along -y direction. The two vertical blue dash lines indicate its sensing region for vector $o$. The $o=(5,1,1,0,1)$ in this case. The blue curve is the design $D$. (b) 2D environment: a robot moves in a 2d plane and lays bricks (red/yellow: in/correct laying). The blue box is its sensing region for $o$, which is a 2d array with size $(2W_s+1)^2$. The green ring is the ground-truth design $D$. (c) 3D environment: a robot moves in a 2d x-y plane and lays bricks (dark blue cubes). Bottom left/right is the 3D/top view. The blue box is similar to 2D. The red ring in the bottom left is the top-most surface of $D$.}
\vspace{-6mm}
\label{fig_example}
\end{figure}

Then, we specialize the above formulation into a suite of progressively more challenging tasks which varies in the setup of grid state $G$ and dimension, action $a$, observation $o$, design $D$ type, and obstacle mechanism, as summarized in Table~\ref{tab_env_detail} and Figure~\ref{fig_example}.

\textbf{Dimension.}
We simplify our mobile construction tasks into three dimension 1/2/3D grid world. The robot is restricted to move along x-axis in 1D and x-y plane for 2D and 3D environment.

\textbf{Action.}
The discrete action $a$ is either moving around or dropping a brick at or near its location.

\textbf{Grid state $G$} stores the number of bricks at each grid. 

\textbf{Observation}
$o$ indicates the current region of the environment observed by the robot in a width-limited window and each element of $o$ represents the number of bricks. To provide more information for the agent, we augment the partial observation $o$ with $N_s$ and $N_b$ into the observation $o_{env}=\langle {o, N_s, N_b} \rangle$ of the environment. $N_s$ is the number of moves the robot has already taken, and $N_b$ is the number of bricks already used.

\textbf{Static} \& \textbf{Dynamic.} We have two types of tasks. The static design task requests the agent to build a static shape D which remains the
same for each episode. For the dynamic design task, the ground-truth design D will vary for each episode. For simulating  real world conditions where robots should have access to the design, we append the design $D$ to the environment observation $o_{env}$ for a dynamic design task as a 4-tuple $o_{env}=\langle {o, N_s, N_b,D} \rangle$. The detail shapes defined for each task are shown in the supplementary. 

\begin{wrapfigure}{R}{0.22\textwidth}
\vspace{-10mm}
  \begin{center}
  \subfloat[][Dense] {\includegraphics[width=0.1\textwidth]{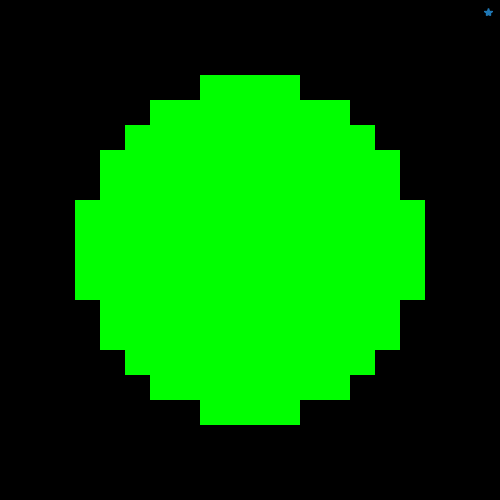}\label{Fig_2D_static_dense}}\hspace{0.005mm}
\subfloat[][Sparse] {\includegraphics[width=0.1\textwidth]{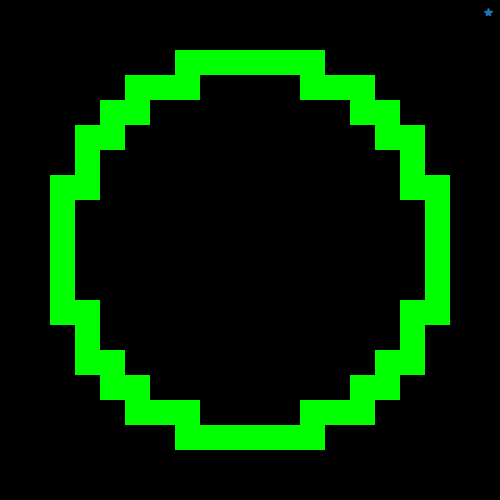}\label{Fig_static_sparse}} 
  \end{center}
  \vspace{-4mm}
  \caption{Example of dense and sparse designs
  \label{fig_dense_and_sparse}}
  \vspace{-5mm}
\end{wrapfigure}
\textbf{Sparse} \& \textbf{Dense.} The term dense means the ground-truth design is a solid shape and sparse means an unfilled shape as shown in Figure~\ref{fig_dense_and_sparse}. 

\textbf{Obstacle.} Different from the 1D and 2D environments where the size of the robot is not considered, the robot size in the 3D environments is considered (occupying 1 grid) so the robot’s motion could be obstructed by the built bricks.

\textbf{Stop criteria.}
Each episode ends when $N_s=N_{smax}$ or $N_b=N_{bmax}$, where the former is the maximum number of steps and the latter the maximum number of bricks (the integrated area of the design $D$ within each episode). We set $N_{smax}$ reasonably large to ensure the design completion. In addition to the same stop criteria as in the 1D/2D cases, the game will stop when the robot is obstructed by the built bricks and cannot move anymore for 3D environment.

\section{Baseline Setup}\label{sec:baseline}
Six state-of-the-art (SOTA) RL baselines and one handcrafted policy are considered in our paper. Here, we fix the environment constants as following: 1) half window size $W_s$ is 2 and width of the environment $W$ is 30 for 1D environments; 2) $W_s$ is 3 and $W$ and $H$ are 20 for 2D and 3D environments. Detailed architecture designs and hyperparameter setups for each algorithm are explained in the supplementary.

\textbf{DQN}.
For the static design tasks, we use an MLP for Q network and three convolutional layers are used to convert the ground-truth $D$ to feature vector for dynamic design tasks. Additionally, instead of only using the current frame, we tried to stack ten frames of historical observations in the replay buffer. This is similar to how the original DQN handles history information for Atari games~\citep{mnih2013playing}, but no significant difference was found.

\textbf{DRQN}.
As for the DRQN \citep{hausknecht2015deep}, we simply add one recurrent LSTM layer to the Q network used in the DQN. 

\textbf{DRQN+Hindsight}. We augment the DRQN baseline with hindsight experience replay~\citep{andrychowicz2017hindsight}. At the end of each training episode, the transitions $\langle {o_{env}^t, a^t, \mathcal{R}(s^t,a^t;D),o_{env}^{t+1}} \rangle$ of each time step $t$ will be relabeled as $\langle {o_{env}^t, a^t, \mathcal{R}(s^t,a^t;G^T),o_{env}^{t+1}} \rangle$, where we change the $D$ to the grid state $G$ at terminate step T. Both of these two transitions will be stored into the replay buffer for optimizing the Q-network.

\textbf{PPO}.
To benchmark PPO in discrete setting, we use the Stable Baselines implementation \citep{stable-baselines}.  

\textbf{Rainbow}.
As for the extended version of DQN, we also include the Rainbow as our baseline. For the Rainbow implementation, we used 3 noisy hidden layers \citep{DBLP:journals/corr/FortunatoAPMOGM17} for the Q network.  

\textbf{SAC}.
 We use the implementation of \citep{christodoulou2019soft} for SAC in discrete setting which has automatic tuning mechanism for entropy hyperparameters.
 
\textbf{Handcrafted policy.}
Besides these RL baselines, we also consider a handcrafted policy with basic localization and planning modules. The localization module borrows the idea of SLAM which uses the common features between successive frames to localize the robot. For the planning modules, we simply let the robot move to the nearest empty grid which should be built. 

\textbf{Human}.
For assessing human performance, we made a simple GUI game (details and video samples of the game in supplementary material) that allows a human player to attempt our tasks in identical environments with the same limitations of partial-observability and step size uncertainty. They act as the agent, using the ARROW keys to maneuver, and SPACE to drop a brick.
\section{Experiments}\label{sec:exp}
In this section, we present testing results of all the baselines on our mobile construction task. The evaluation metric and reward function are defined as following. 

\textbf{Evaluation metric:} we use the IoU score as our evaluation criteria which is measured between the terminal grid state $G^T$ and the ground-truth design $D$. Here, we use IoU as our evaluation criteria because it is more straightforward for us to evaluate the quality of the built structure to the ground-truth design $D$. The IoU is defined as 
\begin{equation}
IoU= \frac{G^T\cap D}{G^T\cup D}= \sum_{l\in\mathcal{L}}\frac{\min( G^T(l),D(l))}{\max(G^T(l),D(l))}.
    \label{eq_IoU}
\end{equation}

\textbf{Reward function.} The reward function $\mathcal{R}$ is for 1D environment is designed as: 
\begin{equation}
\mathcal{R} = 
    \begin{cases}
    10 & \mbox{drops brick on } D, \\
    1 & \mbox{drops brick below } D, \\
    -1 & \mbox{drops brick over } D, \\
    0 & \mbox{moves.}
    \end{cases} 
    \label{reward_1d}
\end{equation}
For the 2D/3D environments, the reward function $\mathcal{R}$ is:
\begin{equation}
\mathcal{R} = 
    \begin{cases}
    5 & \mbox{drops brick at the correct position,}\\
    0 & \mbox{drops brick at the wrong position,}\\
    0 & \mbox{moves.} \\
    \end{cases} 
    \label{reward_2d}
\end{equation}
Note that in 2D environment we tried to add a similar penalty as in equation~\eqref{reward_1d} for incorrect printing, but we find this will lead to the more frequent action of moving instead of printing because of the higher chance of negative rewards of the latter (especially for sparse plans). So we removed the penalty for better performance.



For the static design tasks in 1D/2D/3D, we test the trained agent for 500 times for each task and record the average and min IoU score among 500 tests. For the dynamic design tasks, the ground-truth design is randomly generated during the training phase. In order to make our test results comparable among all the baselines, we define 10 groups of $\sin$ function curves for 1D dynamic and 10 groups of triangles for 2D/3D dynamic task. All algorithms are tested on each group of designs for 200 times and average and min IoU score among 2,000 tests are recorded.

\begin{figure*}[h]
    \centering
    \includegraphics[width=1\linewidth]{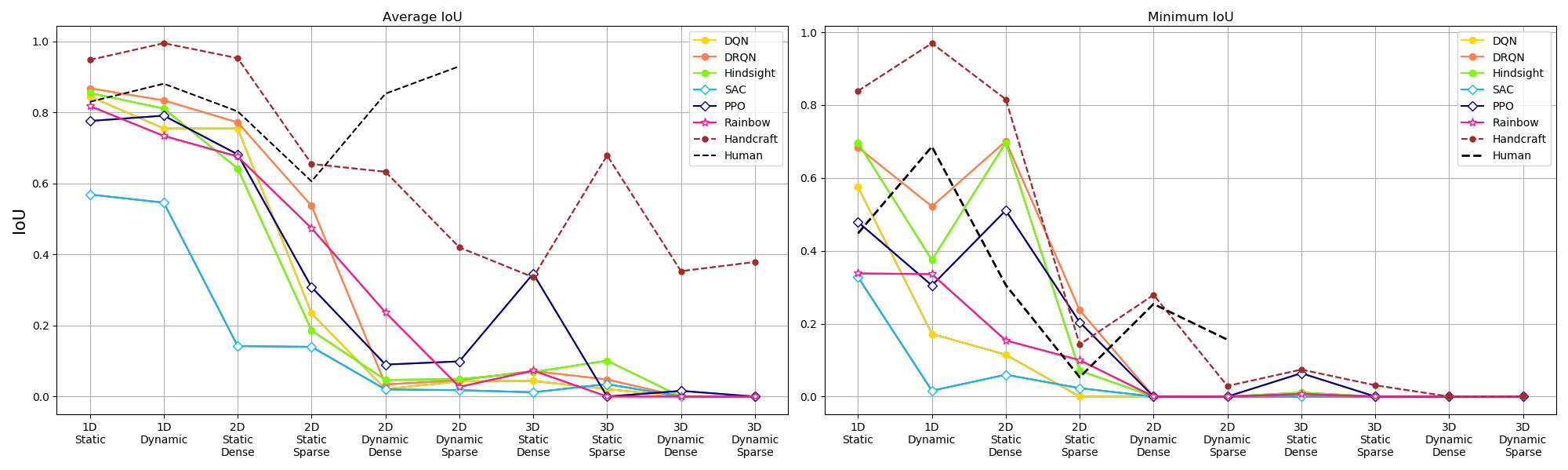}
    \caption{Benchmark results for all baselines, including human baseline: average IoU(left) and minimum IoU(right). Human data of 3D environment is not collected, because it is time-consuming for human to play one game.}
    \label{fig_experiment_results_curve}
    \vspace{-3mm}
\end{figure*}

\begin{figure*}[hbt!]
    \centering
    \includegraphics[width=1\linewidth]{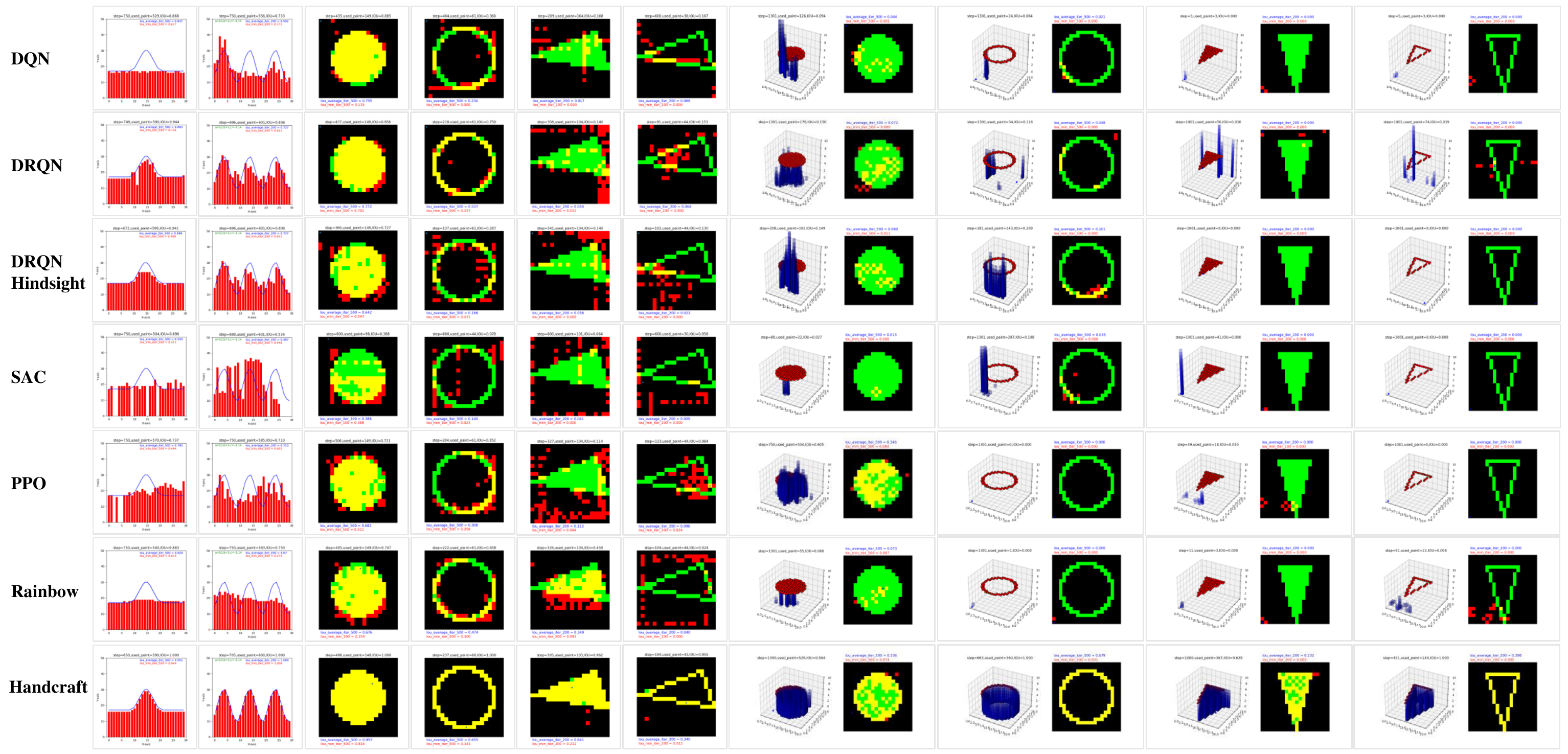}
    \caption{The best testing results of baselines on all tasks. More result videos in the supplementary.}
    \label{fig_best_results}
\vspace{-5mm}
\end{figure*}

\subsection{Benchmark results}
Figure~\ref{fig_experiment_results_curve} summarizes the quantitative results of all baselines on mobile construction tasks. In general, we see that the performance of each methods drops as the dimension of the environment increases. The handcrafted policy outperforms RL baselines in most of the tasks, especially achieving almost IoU 1.0 for 1D tasks. This suggests that this explicitly designed localization and planning method is more effective than model-free RL algorithms. However, as mentioned previously, in the 3D environment, the volume of the agent and obstacle is considered, which makes the tasks substantially more difficult. From the results, most of the deep RL baselines perform rather poorly on the 3D tasks and do not seem to be able to learn any useful policies in the 3D dynamic design tasks. This suggests that the 3D tasks are especially challenging for SOTA RL algorithms and even for handcrafted policy (below IoU 0.4 for 3D dynamic task). We did an ablation study in Section~\ref{sec:ablation} to explore the influence of obstacles on performance. 

From the results, we can also see that the robot has more difficulties performing on dynamic design tasks than on static ones. We posit that the agent lacks the capacity to learn dynamic inputs. DRQN is expected to work best in most of the tasks because it can learn more from a long period of history. Interestingly, the HER does not appear to help noticeably for our tasks. Following, we will discuss the results of different tasks in detail. 

\textbf{1D tasks:} handcrafted policy outperforms other RL baselines on both static (IoU 0.948) and dynamic (IoU 0.995) tasks (details in supplementary Table 1). DRQN has the best performance among all deep RL baselines. For the Gaussian function curve, only Handcrafted policy, DRQN and DRQN+Hindsight could learn a general shape. Other algorithms only managed to build a flatten shape (See first column of Figure~\ref{fig_best_results}). 

\textbf{2D tasks:} similarly, handcrafted policy also achieves best IOU among all baselines. The scores drops significantly from dense to sparse tasks, which indicates the sparse designs are more difficult for handcrafted policy. DRQN has the highest IoU among all RL baselines in the static design tasks.  However, the scores drop dramatically, even close to 0, from static to dynamic plans, which indicates that 2D dynamic designs become much harder for all RL methods.   

\textbf{3D tasks:} we have similarly bad performance in both static and dynamic design tasks. From the last two columns of Figure~\ref{fig_best_results}, we can see that most of the methods can learn to only move in the plane without building any bricks. We hypothesize that the agent learns to only move instead of build bricks in order not to obstruct itself. Handcrafted policy can only achieve below IoU 0.4 for 3D dynamic tasks, which indicates the obstacle is also difficult even for this explicitly designed method. 

\textbf{Handcrafted policy}. Although this handcrafted method performs seemingly well on our 1D tasks, one may not draw a hasty conclusion that this must be the ultimate direction for solving mobile construction.
Because the good performance of this method relies on the prior knowledge and the hyperparameters of the 1D environment: in the current 1D setting, our simple localization has a very low error rate due to the observation window size and the maximum step size, which is not necessarily the case in reality.
Moreover, from Figure~\ref{fig_experiment_results_curve}, we can see that the obstacle mechanism, the sparse designs, and the environment uncertainty in 2D and 3D are still challenging for this method. We believe learning-based methods have a good potential to adaptively address these variations, while solving them one by one via different handcrafted methods is less effective.

\textbf{Human baseline:} we collected 30 groups of human test data, totaling 490 episodes played, and report the average and min IoU showed in the Figure~\ref{fig_experiment_results_curve}. From the feedback, we found that human could learn effective policies (building landmarks to help localization, see our supplementary) very efficiently in at most a few hours,  which is much less than training an RL model. Interestingly, human performed much better on 2D dynamic design tasks.

\subsection{Ablation study}\label{sec:ablation}
We performed ablation studies to comprehensively analyze the reasons associated with the poor baseline performances on static tasks in 2D and 3D. We identify four potential challenges: the obstacles in 3D environments, the lack of localization information, the step size $d$ uncertainty, and the lack of landmarks in the environment for localization. We use DRQN results as the basis of this ablation study and all the ablation experiments were conducted using the same setup described in Section~\ref{sec:baseline}. We use the same test criteria as before. We also conduct a rainbow method ablation study (see supplementary).

\begin{table}[H]
\centering
\scalebox{0.60}{
\begin{tabular}{c|c|ccccc|cc}
\hline
\multirow{2}{*}{Shape} & \multirow{2}{*}{IoU} & \multicolumn{5}{c|}{2D}          & \multicolumn{2}{c}{3D} \\ \cline{3-9} 
 &
   &
  \multicolumn{1}{c|}{\textbf{DRQN}} &
  \multicolumn{1}{c|}{+GPS($\uparrow$)} &
  \multicolumn{1}{c|}{-Uncertainty($\uparrow$)} &
  \multicolumn{1}{c|}{+Obstacle($\downarrow$)} &
  \multicolumn{1}{c|}{+Landmark($\uparrow$)} &
  \multicolumn{1}{c|}{\textbf{DRQN}} &
  -Obstacle($\uparrow$) \\ \hline
\multirow{2}{*}{Dense}       & Avg                  & \textbf{0.772} & +0.079 & +0.073 &   -0.288  & +0.200 & \textbf{0.072}     & +0.727     \\
                             & Min                  & \textbf{0.701} & +0.050 & +0.144 &   -0.568  & +0.179 & \textbf{0}         & +0.714     \\ \hline
\multirow{2}{*}{Sparse}      & Avg                  & \textbf{0.537} & +0.403 & +0.239 &   -0.305  & +0.407 & \textbf{0.048}     & +0.145     \\
                             & Min                  & \textbf{0.237} & +0.149 & +0.539 &   -0.200  & +0.513 & \textbf{0}         & 0          \\ \hline
\end{tabular}}
\caption{Quantitative ablation study results. $\uparrow$/$\downarrow$: expecting better/worse performance.}
\label{tab_ablation_study_results}
\vspace{-3mm}
\end{table}

\begin{figure}[H]

  \centering
  \scalebox{0.73}{
    \includegraphics[width=0.65\textwidth]{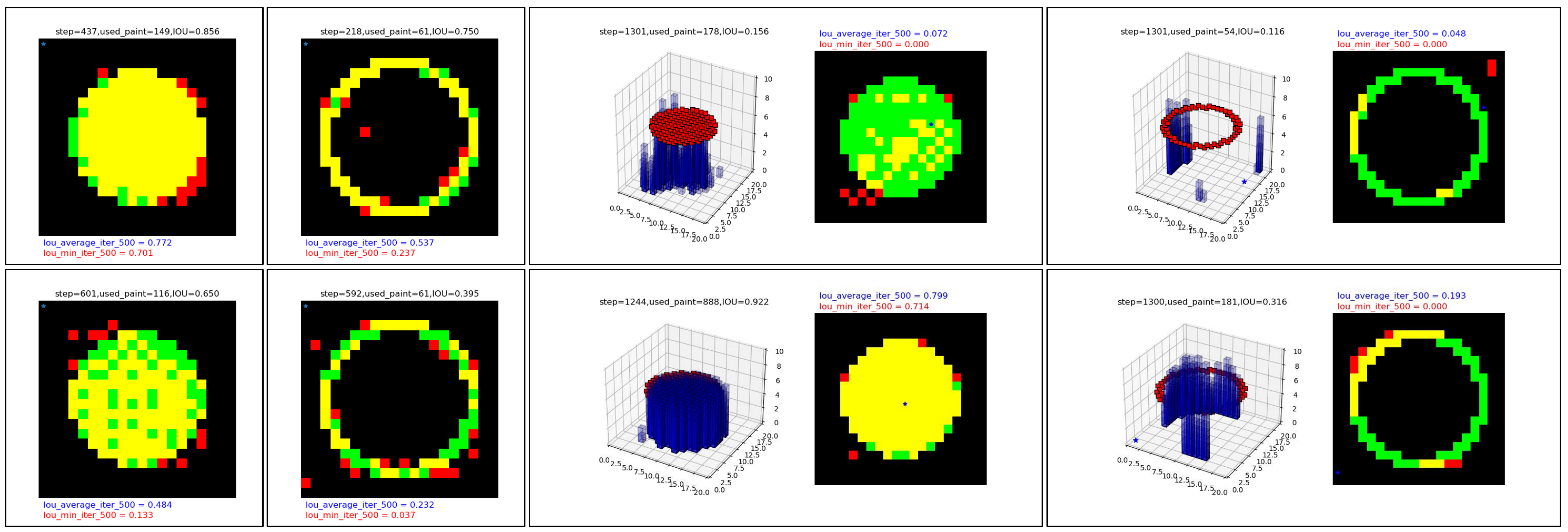}}
 
  \caption{Influences of obstacles. Top rows show the baseline results on 2D and 3D static tasks. The first two images of bottom row show results of adding obstacle to 2D world. Compared with baseline, the performance drops notably. The last two images of bottom row show results of removing obstacle from 3D world. The performance increases significantly, especially for 3D dense design task.      
  \label{fig_remove_obstacle}}
  \vspace{-2mm}
\end{figure}

\textbf{Obstacles.}
To explore the influence of the obstacles, we did two experiments: adding the obstacle mechanism in a 2D task and removing it from the 3D one (Figure~\ref{fig_remove_obstacle}). As showed in Table~\ref{tab_ablation_study_results}, the performance of the 3D task increases significantly from 0.072 to 0.727 for dense designs. Similarly, when we add obstacles into 2D, the IoU score drops by more than $40\%$. These results suggest that the presence of obstacles is an important reason for the poor performances on 3D tasks. 

\begin{figure}[htbp]
\vspace{-3.7mm}
    \centering
    \subfloat[a][Add GPS] {\includegraphics[width=0.235\textwidth]{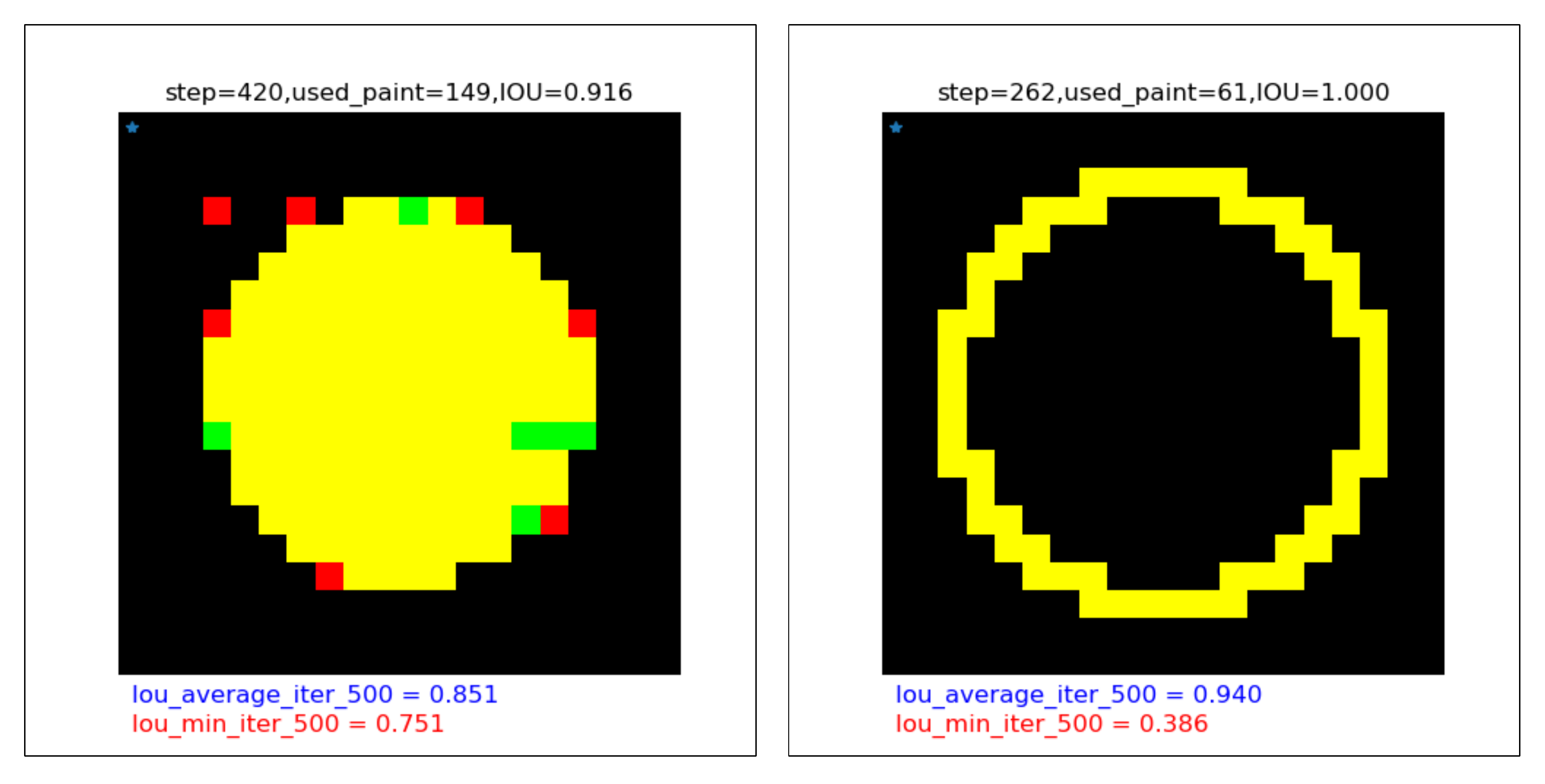}\label{fig_ablation_GPS}} \hspace{0.01mm}
\subfloat[b][Fixed step size] {\includegraphics[width=0.235\textwidth]{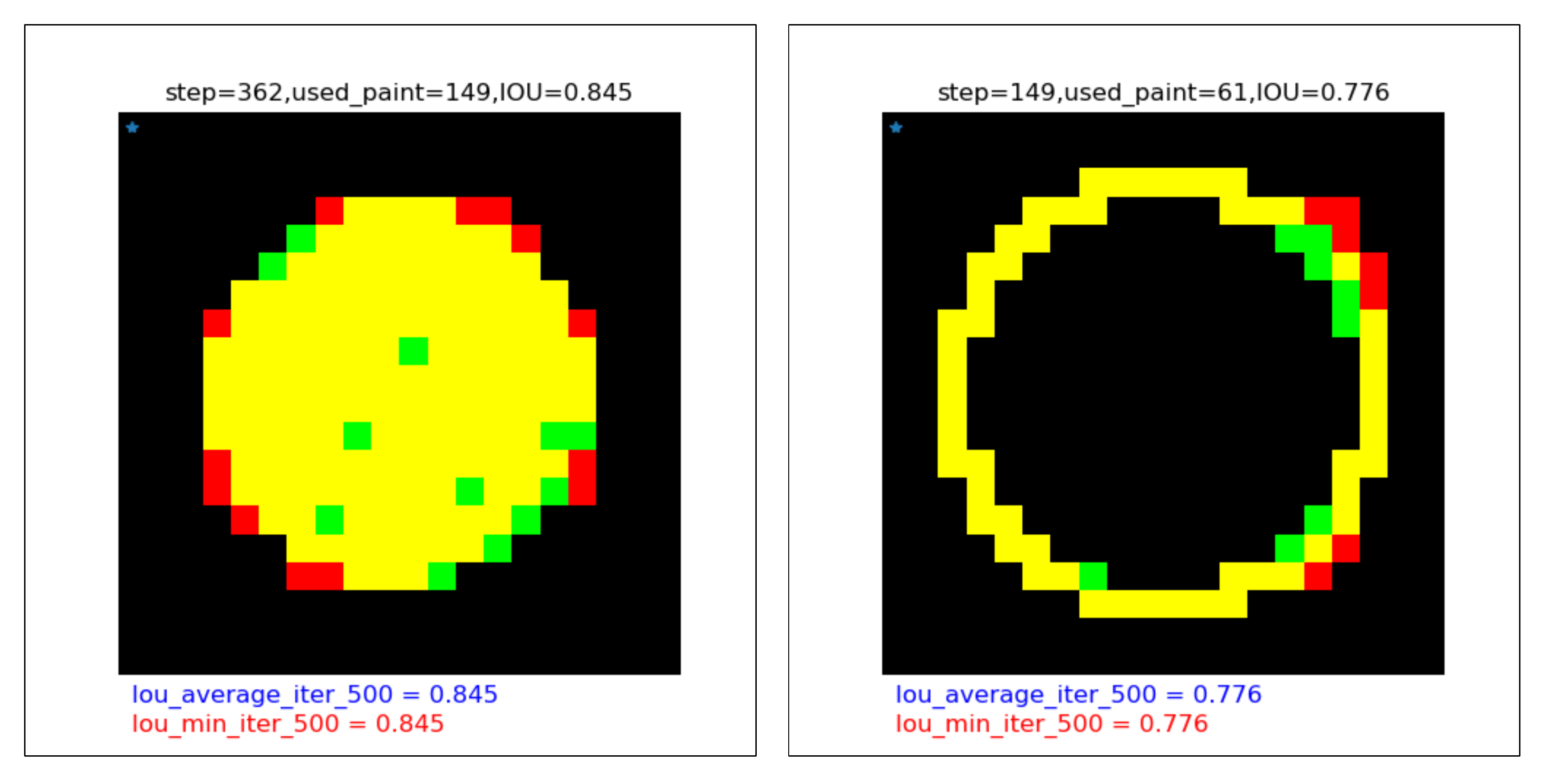}\label{fig_ablation_step}}
\vspace{-2mm}
    \caption{Influence of localization and environment uncertainty. Either adding GPS in (a) or removing step size uncertainty in (b) largely improves DRQN on sparse design tasks compared with the first two columns on the top row of Figure~\ref{fig_remove_obstacle}.}
    \label{fig_ablation}
\vspace{-5mm}
\end{figure}

\textbf{Localization.}
Our mobile construction tasks are intrinsically partially observable and good localization of the robot in its environment becomes a key challenge. The agents are expected to learn implicitly how to do localization with a limited sensing region (partial observation $o$) of the environment. We hypothesize here that the observed low baseline performances in 2D and 3D worlds can further be explained by the fact that the agents lack the capability of localizing precisely. To test this, we provided ground truth location state information $l(x,y)$ for DRQN on 2D static tasks. From Table~\ref{tab_ablation_study_results}, we can see that with the help of position information, the performance increases, especially in 2D sparse task (see Figure~\ref{fig_ablation_GPS}).

\textbf{Environment uncertainty.}
Besides the limited sensing range, the random step size could be another reason for the poor performance of baselines. Therefore, we conduct an experiment in which the step size uncertainty is removed. From Table~\ref{tab_ablation_study_results}, we can see that the IoU increases by more than $40\%$, when compared to the baseline on sparse design (see Figure~\ref{fig_ablation_step}).   

\begin{figure}[htbp]
    \centering
    \subfloat[a][Initial environment] {\includegraphics[width=0.235\textwidth]{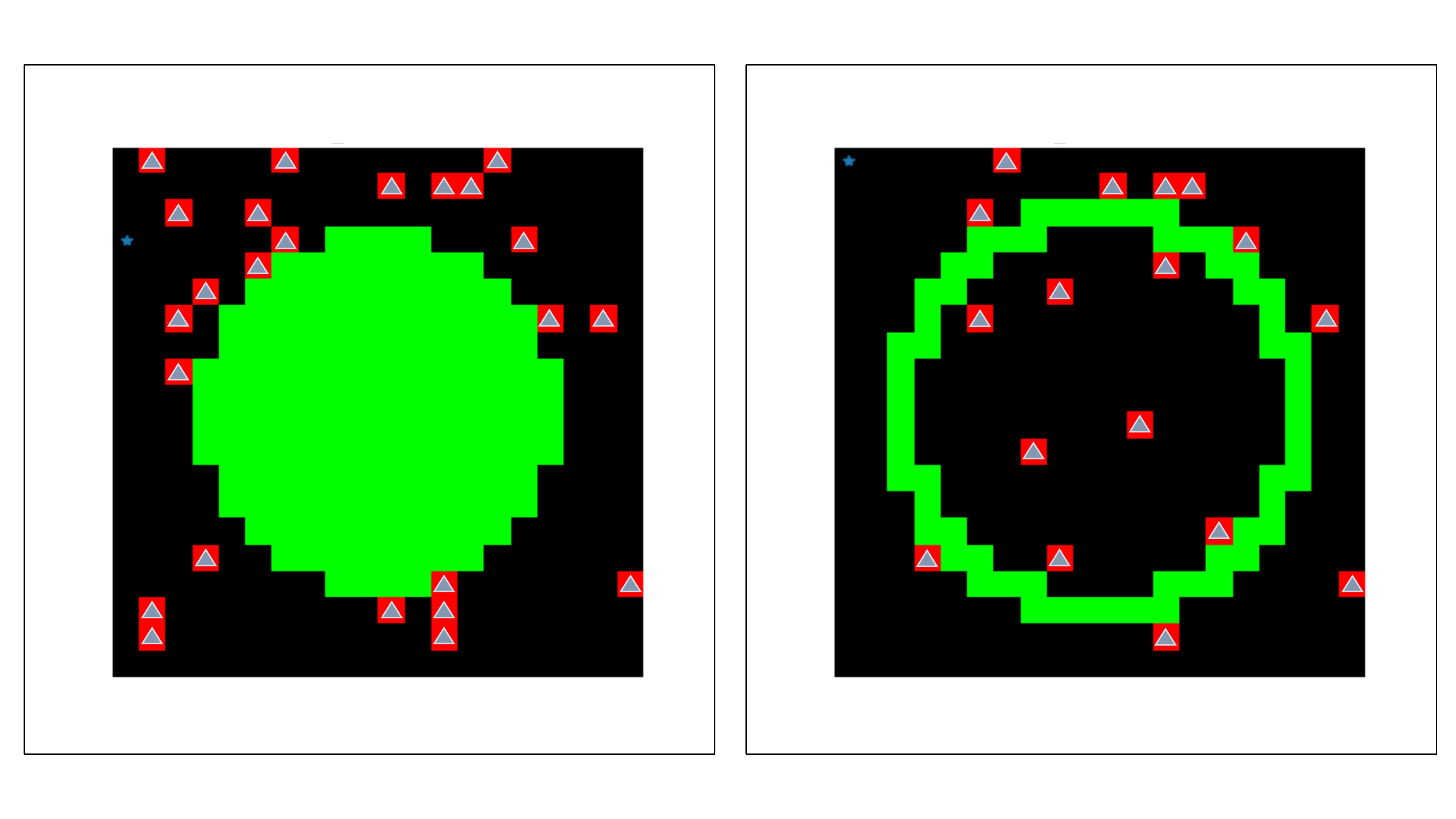}\label{fig_initial_landmark}} \hspace{1mm}
\subfloat[b][Add landmark results] {\includegraphics[width=0.235\textwidth]{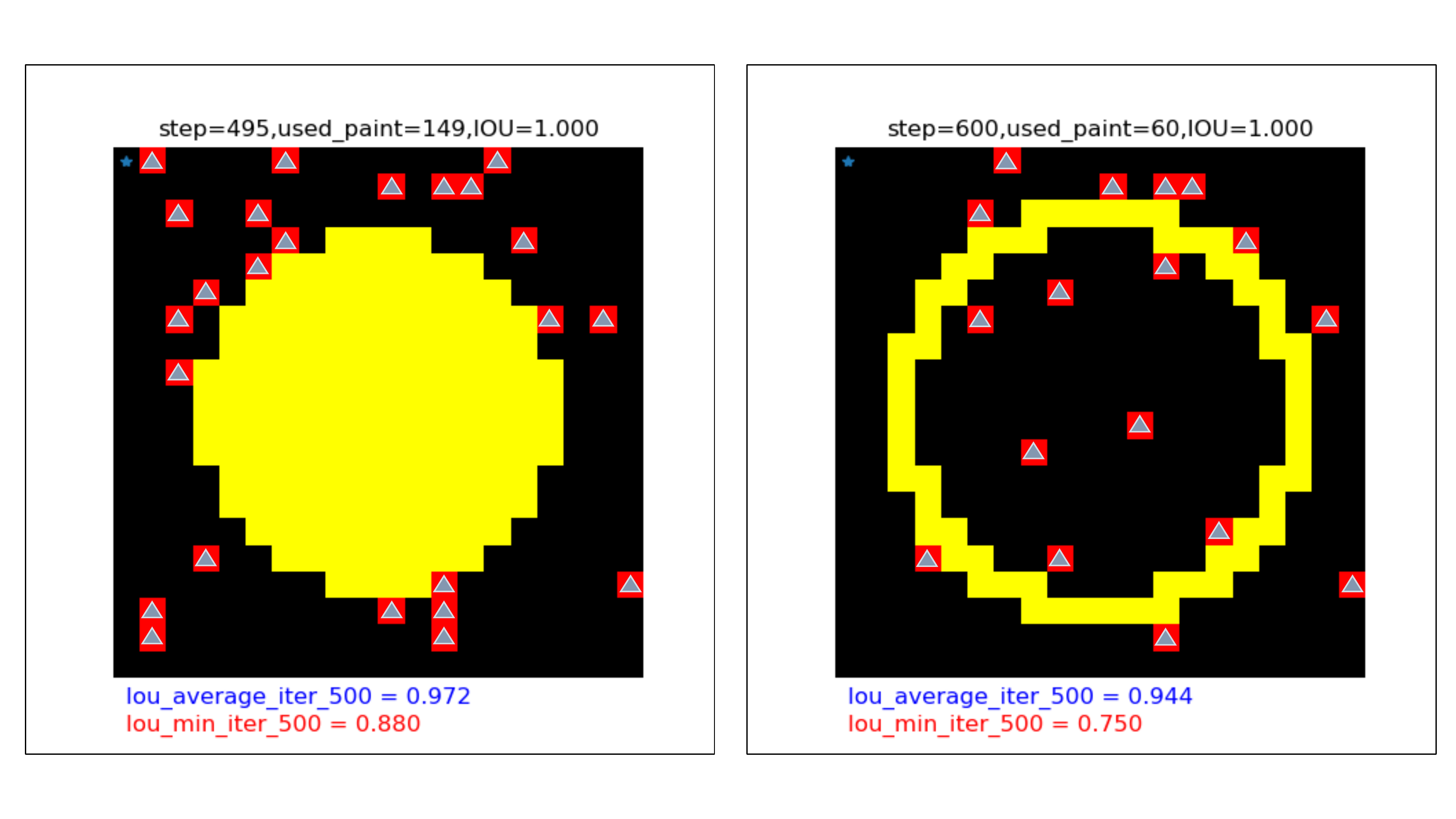}\label{fig_2D_landmark}}
\vspace{-2mm}
    \caption{Influence of landmarks. (a) We randomly add landmarks in initial environment (marked as gray triangle). (b) Compared with baseline, adding landmarks improves DRQN obviously on both dense and sparse design tasks.}
    \label{fig_landmark_ablation}
\vspace{-5mm}
\end{figure}

\textbf{Landmarks.} An empty initial environment lacks landmarks widely used in SLAM methods for robot to localize itself. This could be another potential challenge of mobile construction tasks. We did an experiment: randomly adding some landmarks in the initial environment. From Table~\ref{tab_ablation_study_results}, we can see that IoU increases dramatically both on dense and sparse tasks (see Figure~\ref{fig_landmark_ablation}).  

\vspace{-7mm}
\section{Conclusion and Future Work}
To stimulate the joint effort of robot  localization, planning, and deep RL research, we proposed a suite of mobile construction tasks, where we benchmarked the performance of SOTA deep RL algorithms and a handcrafted policy with basic localization and planning. Our results show that simultaneously navigating a robot while changing the environment is very challenging for those methods. Meanwhile, we also plan to further extend our mobile construction task suite with more features such as allowing design changes within an episode, explicitly adding a landmark placement action, physics-based simulation in continuous worlds, and multi-agent mobile construction.

\section*{Acknowledgment}
The research is supported by NSF CPS program under
CMMI-1932187. The authors gratefully thank
our human test participants and the helpful comments from
Bolei Zhou, Zhen Liu, and
the anonymous reviewers, and also Congcong Wen for paper revision.

\bibliographystyle{IEEEtran}
\bibliography{egbib}
\section*{Appendix}
\setcounter{section}{0}
\section{Baseline setup}
In this section, we state the detail architecture designs and hyperparameter setups for each algorithm.

\textbf{DQN.}
For the static design tasks, we use an MLP with three hidden layers containing [64,128,128] nodes with ReLU activation function for each layer. Three convolutional layers with kernel size of 3, and ReLU activation functions are used to convert the ground-truth $D$ to feature vector for dynamic design tasks. We train DQN on each task for 3,000 episodes. Batch size is 2,000, and replay buffer size 50,000. For tuning the learning rate, we try a relatively small value 1e-8 to make sure convergence initially and gradually increase it until finding a proper value.  Additionally, instead of only using the current frame, we try to stack ten frames of historical observations in the replay buffer. This is similar to how the original DQN handles history information for Atari games~\citep{mnih2013playing}, but no significant difference was found.

\textbf{DRQN.}
The hidden state dimension is 256 for LSTM layer for all tasks. We train it for 10,000 episodes with batch size of 64 and replay memory size of 1,000. 

\textbf{DRQN+Hindsight.}  
We train this DRQN+Hindsight for 10,000 episodes with batch size of 64 and replay memory size of 1,000.

\textbf{PPO.}
 We train PPO for 10 million time steps with a shared network of 3 layers of 512 neurons with tanh activation function. For the hyperparameters, we use the 1D static environment to tune the learning rate, the batch size, the number of minibatches size, and the clipping threshold. We found that the most sensitive parameters were the batch size and the minibatch size and chose the following values: \num{1e5} for the batch size, \num{1e2} for the number of minibatches, \num{2.5e-4} for the learning rate and $0.1$ for the clipping threshold. 

\textbf{Rainbow.}
For the Rainbow implementation, we use 3 noisy hidden layers \citep{DBLP:journals/corr/FortunatoAPMOGM17} with 128 nodes in each layer, and ReLU nonlinear activation functions. Rainbow has a large set of hyperparameters, as each of the six components adds additional hyperparameters.  We used those suggested in \citep{hessel2017rainbow} as a starting point, but they led to poor results on our specific task and environment designs.  As a grid search over such a large hyperparameter space was impractical, we used a random search approach.  Based on empirical results, the algorithm was most sensitive to learning rate, as well as $V_{min}$, $V_{max}$, and $n_{atoms}$, which define the value distribution support predicted by the distributional Q-network.  Generally, a $V_{min}$ value of $-5$, $V_{max}$ of $35$, and $n_{atoms} = 101$ provided stable performance, and were chosen heuristically based on an approximate range of discounted rewards possible in our environments. We used a prioritized experience replay buffer of size \num{1e4}, with priority exponent $\omega$ of $0.5$, and a starting importance sampling exponent $\beta$ of $0.4$.  Additionally, we used multi-step returns with $n = 3$, and noisy network $\sigma_0$ = $0.1$. Finally, we used a learning rate of \num{5e-5} for 1D and 2D, and \num{1e-4} for 3D environments.

 \textbf{SAC.} We use a learning rate \num{3e-4} for target networks for most plans. We use ReLU activation functions for the hidden layers and Softmax for the final layer of the policy network. We use interpolation factor $\tau$=\num{5e-3} for target networks and the start steps before running the real policy is 400 with mini batch size 64. We first search the main hyperparameters based on 1D static case. Next, we use different network architectures for relatively complex 2D and 3D cases with the same hyperparameters as 1D static case. For 1D environment, we use 2 hidden layers with 64 nodes each for both actor and critic networks. For 2D environment, we use 3 hidden layers with 512 nodes each for the dynamic design and 3 hidden layers containing [64, 128, 64] nodes for the static design. A 5-layer network architecture containing [64,128,256,128,64] nodes is applied in the 3D cases.

\textbf{Handcrafted Policy.} The detail handcrafted algorithm is shown in Algorithm~\ref{2d_slam}. The algorithm is divided into two sub-modules, localization  and planning. Localization receives current location $l_t$, current observation $o_{t}$ and next observation $o_{t+1}$ to determine next position $l_{t+1}$ by finding the common features from two successive observations $o_{t}$ and $o_{t+1}$. We borrow this idea from visual SLAM which uses similar mechanism to localize. The planing function will find  all candidate target location $\mathcal{L}_{\text{candidate}}$ for next step by comparing the current observation with the ground truth design $D$. Based on these candidate positions, the agent will decide whether to move to nearest candidate position or build at current position. The priority action space ${A}_{\text{prior}}$ is used to decide the action $a$ when the robot does not receive any specific action commend from the nearest planing policy. This priority action will change when robot touches the boundary of the environment. This mechanism helps robot explore more spaces of the unknown environment.

\begin{figure}[h]
\vspace{-8mm}
\centering
  \subfloat[][Training (1D).] {\includegraphics[width=0.252\textwidth]{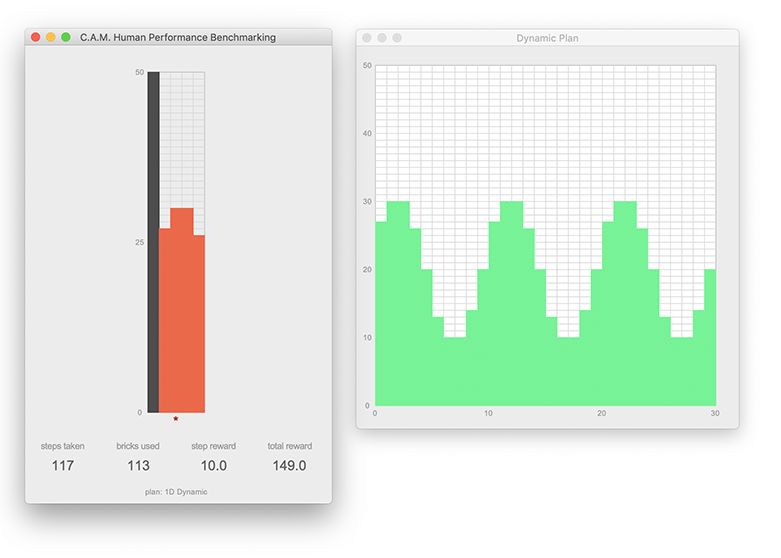}\label{Fig.sub.GUI1}}
\subfloat[][Evaluation (2D).] {\includegraphics[width=0.192\textwidth]{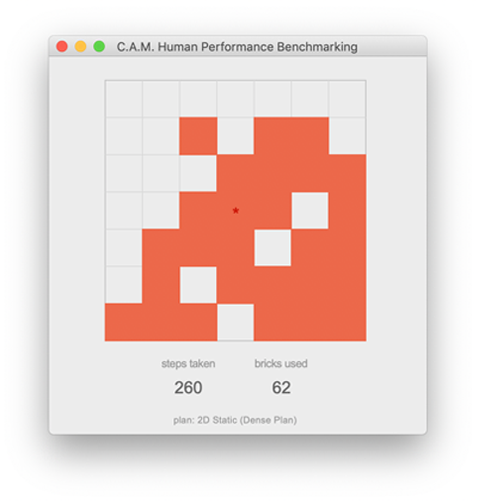}\label{Fig.sub.GUI2}}\quad  
  \vspace{-2mm}
  \caption{Game GUI for measuring human performance.  (a) 1D Dynamic design: user can see the ground truth design in dynamic environments.  (b) 2D Static design in \textit{evaluation mode}: only step and brick count are shown, while rewards are hidden.     
  \label{fig_game_GUI}}
  \vspace{-4mm}
\end{figure}

\textbf{Human.} The human baseline GUI game is shown in Figure~\ref{fig_game_GUI}. In static environments, human players are required to complete the task without the access to the ground-truth design, while in dynamic environments, they can reference the current ground truth design. Additionally, players can toggle between training or evaluation mode.  In training mode, they can view per-step reward as well as their cumulative reward over the episode, whereas in evaluation mode, they can only see the number of bricks used and steps taken.  For each episode played, players reported their episode-IoU.

\begin{algorithm}[H] 
\caption{Handcrafted Policy}\label{2d_slam}
\begin{algorithmic}[1]
\Statex
\Function{Main}{}
\State Initialize environment and get initial observation $o_{t=0}$
\State Get initial location $l_{t=0}$
\State Initialize priority action space $\mathcal{A}_{\text{prior}}$

\For{$t\gets 0,N_{smax}$}
    \State $a, \mathcal{A}_{\text{prior}}$=\Call {Planning}{$l_t,o_t,\mathcal{A}_{\text{prior}}$}
    \State Execute $a$ and obtain next observation $o_{t+1}$, reward, and terminal sign Done from environment.
    \State $l_{t+1}$ = \Call{Localization}{$l_{t},o_t,o_{t+1},a$}
    
    \If{Done}
        \State Break
    \EndIf
\EndFor
\EndFunction
\Statex
\Function{Localization}{$l_{t}$,$o_t$,$o_{t+1}$,$a$}
\State Common Feature = \Call{FeatureMatching}{$o_t,o_{t+1}$}
    \If{Common Feature is empty \textbf{or} $o_t=o_{t+1}$}
        \State Determine $l_{t+1}$ using Odometry, here we assume step size is always 1.
        \Return $l_{t+1}$
    \Else 
        \State Determine step size using Common Feature.
        \State Calculate $l_{t+1}$ using step size and $a$.
        \Return $l_{t+1}$
    \EndIf
\EndFunction
\Statex
\Function{Planning}{$l_t,o_t,\mathcal{A}_{\text{prior}}$}
\State $\mathcal{L}_{\text{candidate}}$ = \Call{Compare}{$o_t,D(l_t)$}
\State Update $\mathcal{A}_{\text{prior}}$ based on the current boundary condition
    \If{$\mathcal{L}_{\text{candidate}}$ is empty}
        \State $a$ is random sampled from $\mathcal{A}_{\text{prior}}$
        \Return $a, \mathcal{A}_{\text{prior}}$
    \Else 
        \State Find nearest location $l_{\text{near}}$ from $\mathcal{L}_{\text{candidate}}$
    \EndIf
    \If{$l_{\text{near}}=l_t$} 
        \Return $a$=build brick, $\mathcal{A}_{\text{prior}}$ 
    \Else 
        \State Determine action $a$ based on the corresponding direction of $l_{\text{near}}$ to $l_t$
        \Return $a, \mathcal{A}_{\text{prior}}$
    \EndIf
\EndFunction
\end{algorithmic}
\end{algorithm}

\section{Benchmark result}
In this section, we show the quantitative results of each baseline on all tasks in Table~\ref{tab_benchmark_results} and the Rainbow method ablation study.

\begin{table}[t]
\centering
\resizebox{\linewidth}{!}{
\begin{tabular}{c|l|llllllllll}
\hline
                                 & \multicolumn{1}{c|}{} & \multicolumn{2}{c|}{1D} & \multicolumn{4}{c|}{2D}       & \multicolumn{4}{c}{3D} \\ \cline{3-12} 
 &
  \multicolumn{1}{c|}{IoU} &
  \multicolumn{1}{c|}{\multirow{2}{*}{Static}} &
  \multicolumn{1}{c|}{\multirow{2}{*}{Dynamic}} &
  \multicolumn{2}{c|}{Static} &
  \multicolumn{2}{c|}{Dynamic} &
  \multicolumn{2}{c|}{Static} &
  \multicolumn{2}{c}{Dynamic} \\ \cline{5-12} 
 &
  \multicolumn{1}{c|}{} &
  \multicolumn{1}{c|}{} &
  \multicolumn{1}{c|}{} &
  \multicolumn{1}{l|}{Dense} &
  \multicolumn{1}{l|}{Sparse} &
  \multicolumn{1}{l|}{Dense} &
  \multicolumn{1}{l|}{Sparse} &
  \multicolumn{1}{l|}{Dense} &
  \multicolumn{1}{l|}{Sparse} &
  \multicolumn{1}{l|}{Dense} &
  Sparse \\ \hline
\multirow{2}{*}{\textbf{Human}}           & Avg                   & 0.83      & 0.881      & 0.803 & 0.606 & \textbf{0.853} & \textbf{0.93} &$-$  & $-$  &$-$   &$-$  \\
                                 & Min                   & 0.448      & 0.686      & 0.306 & 0.053 & 0.254   & 0.156   & $-$  & $-$    & $-$  & $-$ \\ \hline
\multirow{2}{*}{DQN}             & Avg                   & 0.845      & 0.755      & 0.755 & 0.234 & 0.021 & 0.043 & 0.044  & 0.021  &   &  \\
                                 & Min                   & 0.575      & 0.172      & 0.115 & 0 & 0   & 0   & 0.001  & 0    &   &  \\ \hline
\multirow{2}{*}{DRQN}            & Avg                   & 0.868      & 0.834      & 0.772 & 0.537 & 0.034 & 0.046 & 0.072  & 0.048   &   &  \\
                                 & Min                   & 0.684      & 0.522      & 0.701 & 0.237 & 0   & 0   & 0  & 0    &   &  \\ \hline
\multirow{2}{*}{DRQN+Hindsight} & Avg                   & 0.855      & 0.811      & 0.642 & 0.186 & 0.047 & 0.049 & 0.069  & 0.101    &   &  \\
                                 & Min                   & 0.697      & 0.375      & 0.697 & 0.071 & 0   & 0   & 0.011    & 0    &   &  \\ \hline
\multirow{2}{*}{SAC}             & Avg                   & 0.569     & 0.546      &  0.142      & 0.14 & 0.02   & 0.018     & 0.012     & 0.035       &   &  \\
                                 & Min                   & 0.328      & 0.016      &  0.06     & 0.023 & 0     & 0     & 0        &   0     &   &  \\ \hline
\multirow{2}{*}{PPO}             & Avg                   & 0.776      & 0.791      & 0.682 & 0.308 & 0.09  & 0.099      & \textbf{0.346}    &        &  0.016 &   \\
                                 & Min                   & 0.479      & 0.305      & 0.511 & 0.204 & 0  & 0 & 0.064       &    &  0 &  \\ \hline
\multirow{2}{*}{Rainbow}         & Avg                   & 0.818      & 0.734      & 0.676 & 0.474 & 0.237 & 0.027       &  0.073       &        &   &  \\
                                 & Min                   & 0.338          &    0.336        &  0.154     & 0.1      &  0    &  0    &  0.007      &        &   &  \\ \hline
\multirow{2}{*}{Handcraft}       & Avg                   & \textbf{0.948}      & \textbf{0.995}      & \textbf{0.953}  & \textbf{0.655} & 0.633       &  0.42       &0.336        &\textbf{0.679} &\textbf{0.353}   &\textbf{0.379}  \\
                                 & Min                   & 0.839      & 0.970      &  0.816     & 0.143      &  0.279    &  0.029    &  0.074      &0.031        &0   &0  \\ \hline
\end{tabular}%
}
\caption{Benchmark quantitative results. Empty cells indicate the agents failed at the these task without learning any control policy.}
\label{tab_benchmark_results}
\vspace{-2mm}
\end{table}

\begin{figure}[H]
    \begin{center}
\includegraphics[width=0.4\textwidth]{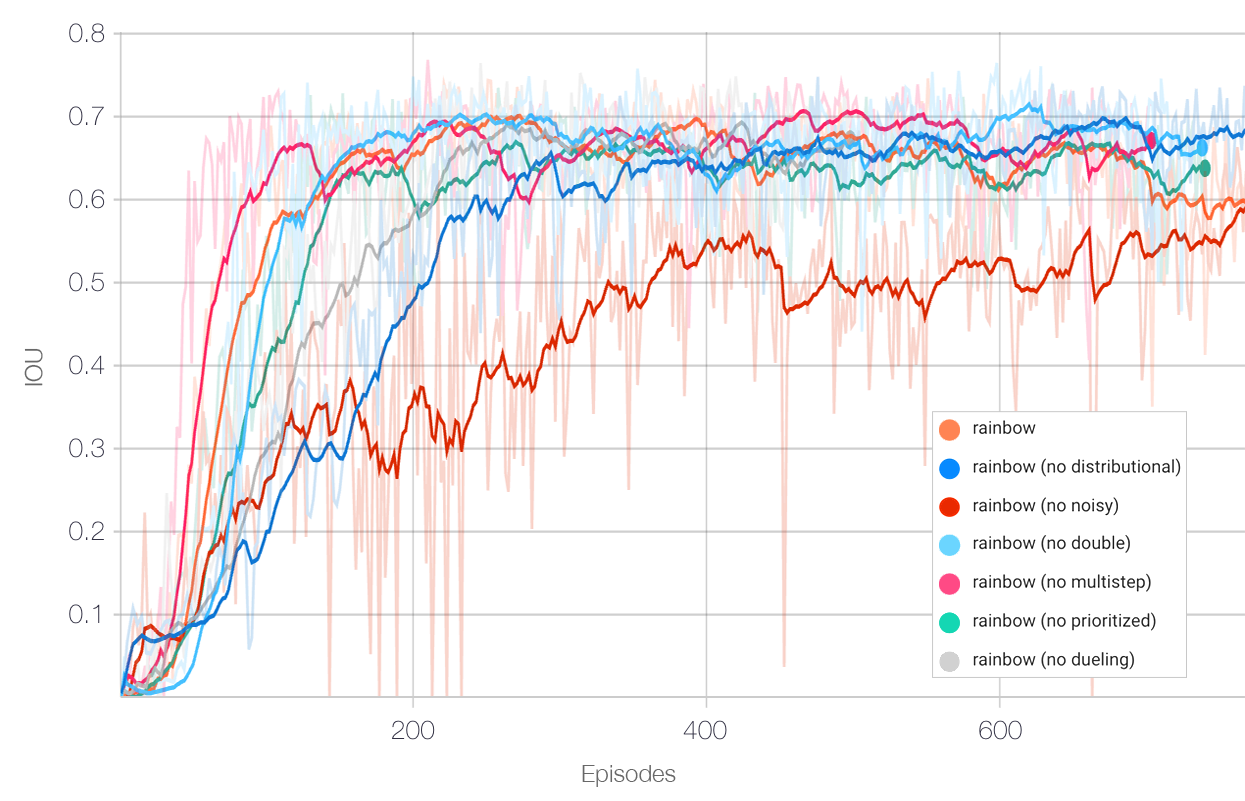}
\caption{Training history of IoU: the full Rainbow algorithm does not always outperform its pruned configurations. 
        \label{fig_rainbow_ablation}}
    \end{center}
\vspace{-5mm}
\end{figure}

\textbf{Rainbow method ablation study.}
We observed similar performance across most tasks when comparing the base DQN algorithm with DQN plus Rainbow. To better understand the effect of each individual extension on the whole, we conducted six separate runs: in each, we removed one extension from the complete Rainbow algorithm, similar to \citep{hessel2017rainbow}. Figure~\ref{fig_rainbow_ablation} displays the IoU running average over training in the 2D Static environment for each of these pruned configurations. Overall, we find that the complete Rainbow algorithm initially learns \textit{faster} than all but one of the pruned configurations, suggesting that the combination of optimizations indeed leads to better sample-efficiency in the learning process. Removing noisy networks (reverting to an epsilon-greedy exploration approach) led to the largest decrease in learning efficiency. Over a longer horizon, the top testing performance plateaus to nearly the same across all configurations. In the case of distributional learning, multi-step learning, and double DQN, removing each individual actually improved top test performance, from 0.676 to 0.685, 0.701, and 0.714 respectively. While this study is not exhaustive, it suggests that the Rainbow algorithm does not easily generalize from the Arcade Learning Environment~\citep{Bellemare_2013} (which it was designed for) to our specific tasks.

\end{document}